\documentclass[10pt,twocolumn,letterpaper]{article}
\usepackage{amsmath,amsfonts,bm}

\def\eqref#1{equation~\ref{#1}}
\def\1{\bm{1}}

\def\rvf{{\mathbf{f}}}

\def\vp{{\bm{p}}}
\def\vq{{\bm{q}}}

\def\vz{{\bm{z}}}

\DeclareMathAlphabet{\mathsfit}{\encodingdefault}{\sfdefault}{m}{sl}
\SetMathAlphabet{\mathsfit}{bold}{\encodingdefault}{\sfdefault}{bx}{n}

\usepackage[pagenumbers]{cvpr} % To force page numbers, e.g. for an arXiv version

\usepackage{graphicx}
\usepackage{amsmath}
\usepackage{amssymb}
\usepackage{booktabs}
\usepackage{times}
\usepackage{epsfig}
\usepackage{bm}
\usepackage{float}
\usepackage{pifont}

\usepackage[breaklinks,pagebackref,bookmarks=false,colorlinks,citecolor=OliveGreen]{hyperref}

\usepackage[capitalize]{cleveref}
\crefname{section}{Sec.}{Secs.}
\Crefname{section}{Section}{Sections}
\Crefname{table}{Table}{Tables}
\crefname{table}{Tab.}{Tabs.}

\usepackage[dvipsnames]{xcolor}
\usepackage[ruled,vlined]{algorithm2e}
\usepackage[symbol]{footmisc}

\makeatletter
\newcommand{\algorithmfootnote}[2][\footnotesize]{%
  \let\old@algocf@finish\@algocf@finish% Store algorithm finish macro
  \def\@algocf@finish{\old@algocf@finish% Update finish macro to insert "footnote"
    \leavevmode\rlap{\begin{minipage}{\linewidth}
    #1#2
    \end{minipage}}%
  }%
}
\makeatother

\definecolor{moco}{RGB}{33, 130, 129}

\SetKwComment{Comment}{\color{moco}\# }{}

\usepackage{makecell}
\usepackage{multirow}

\def\cvprPaperID{1295} % *** Enter the CVPR Paper ID here
\def\confName{CVPR}
\def\confYear{2022}

\begin{document}

\title{Self-distillation with Batch Knowledge Ensembling\\ Improves ImageNet Classification}

\author{
Yixiao Ge$^1$ \quad Xiao Zhang$^1$ \quad Ching Lam Choi$^1$ \quad Ka Chun Cheung$^3$ \quad Peipei Zhao$^5$ \\
\vspace{5pt}
Feng Zhu$^4$ \quad Xiaogang Wang$^1$ \quad Rui Zhao$^4$ \quad Hongsheng Li$^{1,2}$ \\
% $^1$CUHK-SenseTime Joint Laboratory, The Chinese University of Hong Kong\\
$^1$Multimedia Laboratory, The Chinese University of Hong Kong\\
$^2$Centre for Perceptual and Interactive Intelligence (CPII)\\
$^3$NVIDIA\quad
$^4$SenseTime Research\quad
$^5$School of CST, Xidian University\\
\vspace{5pt}
{\tt\small \{yxge@link, hsli@ee\}.cuhk.edu.hk}\\
{\small Project Page: \url{https://geyixiao.com/projects/bake}}
}
% \and
\maketitle

%%%%%%%%% ABSTRACT
% \vspace{-5pt}
\begin{abstract}
\vspace{-5pt}
   The recent studies of knowledge distillation \cite{lan2018one,shen2019meal,tian2019crd,guo2020online} have discovered that ensembling the ``dark knowledge'' from multiple teachers or students contributes to creating better soft targets for training, but at the cost of significantly more computations and/or parameters. In this work, we present BAtch Knowledge Ensembling (BAKE) to produce refined soft targets for anchor images by propagating and ensembling the knowledge of the other samples in the same mini-batch. Specifically, for each sample of interest, the propagation of knowledge is weighted in accordance with the inter-sample affinities, which are estimated on-the-fly with the current network. The propagated knowledge can then be ensembled to form a better soft target for distillation. In this way, our BAKE framework achieves online knowledge ensembling across multiple samples with only a single network. It requires minimal computational and memory overhead compared to existing knowledge ensembling methods. Extensive experiments demonstrate that the lightweight yet effective BAKE consistently boosts the classification performance of various architectures on multiple datasets, \textit{e.g.}, a significant \textbf{+0.7\%} gain of Swin-T on ImageNet with only $+1.5\%$ computational overhead and \textbf{zero} additional parameters. BAKE does not only improve the vanilla baselines, but also surpasses the single-network state-of-the-arts \cite{furlanello2018born,yun2020regularizing,yuan2020revisiting,zhang2019your,xu2019data} on all the benchmarks.
%   \footnote{Code: \url{https://github.com/Pre-release/BAKE}}
\end{abstract}

%%%%%%%%% BODY TEXT
\vspace{-10pt}
\section{Introduction}
\label{sec:introduction}

\begin{figure}[t]
\centering
\includegraphics[width=0.65\linewidth]{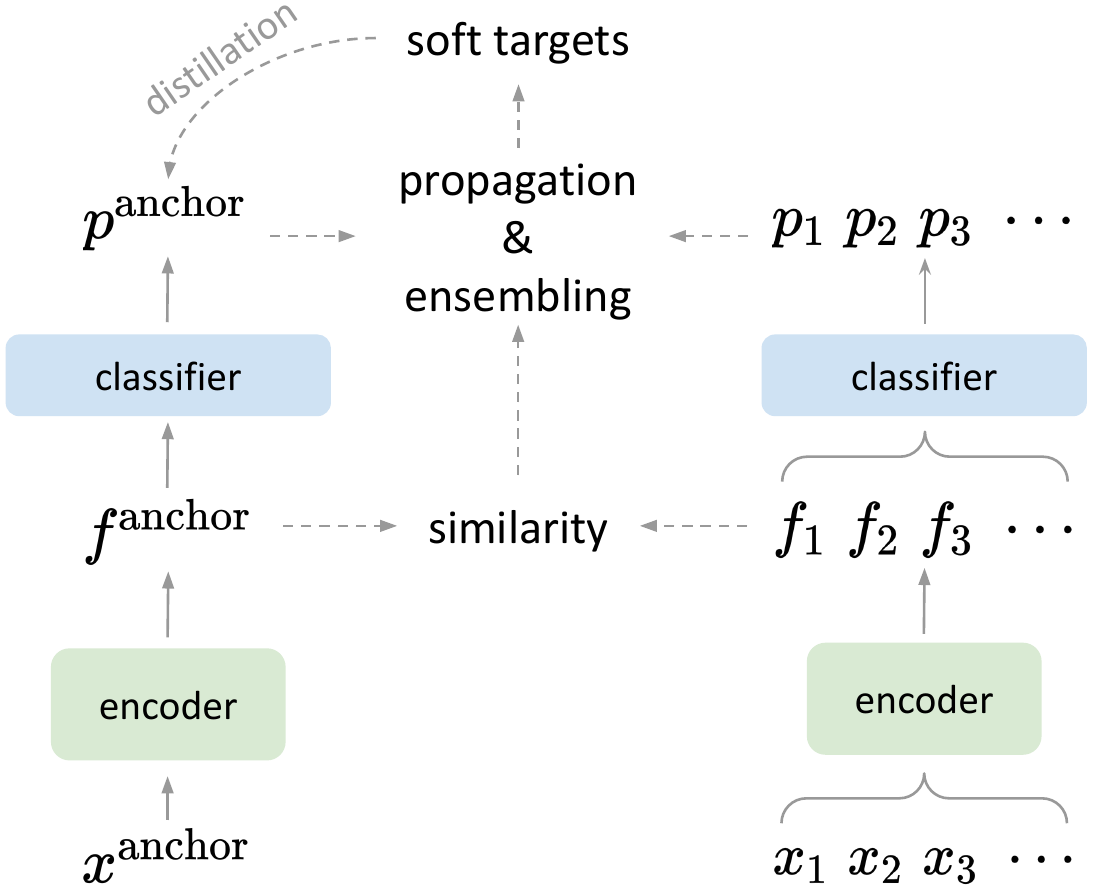}
\vspace{-5pt}
\caption{BAtch Knowledge Ensembling (BAKE) produces soft targets for self-distillation \emph{with a single network} (an encoder and a classifier). For an anchor image $x^\text{anchor}$, the knowledge of the other samples $\{x_1, x_2, x_3, \cdots\}$ in the same batch is weightedly propagated and ensembled to form better soft targets for distillation on-the-fly. Note that $x^\text{anchor}$ and $\{x_1, x_2, x_3, \dots\}$ are fed into the same network. This method enables knowledge ensembling with minimal computational and memory overhead.}
\vspace{-10pt}
\label{fig:intro}
\end{figure}

Deep neural networks have achieved impressive success on computer vision tasks, where image classification \cite{he2016deep,Xie2016,howard2017mobilenets,tan2019efficientnet,zhang2020resnest} is considered as one of the most fundamental tasks given the wide range of applications of its learned representations and their transferability to downstream tasks, 
\textit{e.g.}, detection \cite{fasterrcnn,lin2017feature}, segmentation \cite{zhao2017pyramid,he2017mask}, generation \cite{CycleGAN2017,pix2pix2017,Liu_DivCo,ge2020structured}, retrieval \cite{ge2018fd,ge2020mutual,ge2020self,ge2020selfpaced}, \textit{etc}. 
There is a tremendous number of methods \cite{bagherinezhad2018label,beyer2020we,xie2020self,guo2020online,yun2020regularizing,yuan2020revisiting,yun2021relabel} working on improving the image classification accuracy, especially on the large-scale ImageNet \cite{deng2009imagenet} dataset. 

Recent studies \cite{bagherinezhad2018label,beyer2020we,yun2021relabel} have shed light on the limitations of supervised learning of image classification. 
They have observed that the imperfect learning targets resulted from manually annotated ground-truth labels (one-hot class vectors) turn out to be a key factor that hinders the further improvement of classification accuracy.
Thanks to the great success of knowledge distillation \cite{kd}, 
the soft probability vectors predicted by a teacher network, carrying the learned ``dark knowledge'', 
can serve as informative training supervisions to enhance a student network.
The quality of the teacher's predictions is found critical to the accuracy of the student network.
State-of-the-art methods \cite{lan2018one,shen2019meal,tian2019crd,guo2020online,son2021densely} have found that multiple teachers or students could encode complementary knowledge and their ``ensembled'' soft targets are more robust learning objectives. 
Positive as their results are, they depend on extra networks or branches, undoubtedly increasing the computational and memory cost to a noticeable extent.

To produce high-quality soft targets \emph{with minimal cost}, we introduce a self-distillation method with a novel \textbf{BA}tch \textbf{K}nowledge \textbf{E}nsembling (\textbf{BAKE}) scheme, as illustrated in Figure \ref{fig:intro}.
Rather than ensembling multiple networks or branches to generate the distillation soft targets, our method only adopts a single network. It achieves knowledge ensembling by on-the-fly aggregating the ``dark knowledge'' from different samples within the same mini-batch, yielding better soft targets.
% Note that the way of ``knowledge ensembling'' is not limited to multi-model ensembling, \textit{e.g.}, in semi/self-supervised learning tasks, \cite{meanteacher,he2020momentum,byol} use temporal ensembling with momentum updates to integrate the knowledge of a student model over iterations.

Specifically, given the samples' encoded representations and predictions in a mini-batch, 
we conduct cross-sample knowledge ensembling under the assumption that visually similar samples with close-by representations should encode consistent class-wise predictions. In practice, for each anchor sample, the other samples' predictions (``dark knowledge'') can be weightedly propagated and ensembled to form a soft target. 
The knowledge propagation and ensembling are conducted iteratively until convergence, \textit{i.e.}, the soft targets no longer change.
In order to perform the proposed batch knowledge ensembling at each training iteration efficiently, we adopt approximate inference to estimate the iterative knowledge ensembling results. After properly ensembling the samples' knowledge in the batch, we are able to create refined soft targets for each sample to improve the training of image classification.

We demonstrate the effectiveness of our proposed BAKE via a comprehensive set of evaluations with various architectures and datasets.
On ImageNet \cite{deng2009imagenet} classification, BAKE well boosts the top-1 accuracy of ResNet-50 \cite{he2016deep} by significant \textbf{+1.2\%} gains (76.8\%$\to$78.0\%) with a negligible {3.7\%} computational overhead at training.
We also evaluate our BAKE on vision transformers, \textit{i.e.}, Swin Transformer \cite{liu2021swin}, and boost the state-of-the-art performance by up to \textbf{+0.7\%} improvements (81.3\%$\to$82.0\%).
% in terms of top-1 accuracy on ImageNet.
The network trained by BAKE shows consistent improvements for transfer learning on downstream tasks, including detection \cite{fasterrcnn} and segmentation \cite{he2017mask} on COCO \cite{coco}.
BAKE also improves the classification robustness on perturbed datasets \cite{fgsm,hendrycks2019robustness,hendrycks2021nae} at test time.
Not only improving the accuracy on ImageNet, BAKE also improves the classification accuracies on CIFAR-100 \cite{cifar}, TinyImageNet, CUB-200-2011 \cite{cub}, MIT67 \cite{mit} and Stanford Dogs \cite{dog}.
BAKE substantially outperforms the single-network state-of-the-arts \cite{furlanello2018born,yun2020regularizing,yuan2020revisiting,zhang2019your,xu2019data} on all the benchmarks.

The contributions of our method are three-fold.
(1) We for the first time introduce to produce ensembled soft targets for self-distillation without using multiple networks or additional network branches.
(2) We propose a novel batch knowledge ensembling mechanism to online refine the distillation targets with the cross-sample knowledge, \textit{i.e.}, weightedly aggregating the knowledge from other samples in the same batch.
(3) Our method is simple yet consistently effective on improving classification performances of various networks and datasets with minimal computational overhead and zero additional parameters.

\section{Related Works}

% \vspace{-5pt}
\paragraph{Knowledge distillation.} 
Knowledge distillation \cite{kd} aims to transfer the ``dark'' knowledge learned from a high-capacity teacher network to a student network via soft labels. 
The soft labels can be the class probabilities \cite{furlanello2018born,guo2020online,yun2020regularizing} or the feature representations \cite{tung2019similarity,park2019relational} output by the teacher, containing more complete structured information than the one-hot ground-truth labels. 
The distillation process can be formed by a ``teacher-student'' framework \cite{kd,shen2019meal}, a ``peer-teaching'' framework \cite{lan2018one,zhang2018deep,guo2020online}, or a self-distillation framework \cite{furlanello2018born,zhang2019your,xu2019data,yun2020regularizing,kim2021self}. 
% Our BAKE is mostly related to the last one.
% %
% Traditional methods \cite{kd} adopted the normal ``teacher-student'' framework to distill the knowledge from pre-trained teacher networks to the student network in a two-stage training phase.
% Follow-up works \cite{lan2018one,zhang2018deep,guo2020online} found the traditional way is less efficient, thus introduced a one-stage manner, \textit{i.e.}, performing peer-teaching among a set of simultaneously training students on-the-fly.
% Recent literature further introduced self-distillation methods \cite{furlanello2018born,zhang2019your,xu2019data,yun2020regularizing} to teach a single network using its own knowledge, which are the most flexible to date due to their low computational and memory overhead.
Our BAKE is mostly related to the self-distillation methods, \textit{i.e.}, teaching a single network using its own knowledge.
% which are the most flexible to date due to their low computational and memory overhead.
However, most of them \cite{furlanello2018born,zhang2019your,kim2021self} only considered the knowledge of individual instances, resulting in sub-optimal learning targets.
Recent works introduced to preserve the predictive consistency between intra-image (original \textit{v.s.} perturbed) \cite{xu2019data} or intra-class (images out of the same class) \cite{yun2020regularizing} samples. 
However, they only focused on pairwise images, carrying limited information compared to the ensembled batch knowledge of BAKE.
More importantly, they simply defined the positive pairs using constant instance or class IDs, which may incur false supervisions as their visual features might be actually dissimilar, especially after the random crop augmentation \cite{szegedy2015going}. 

\begin{table}[t]
\footnotesize
\begin{center}
	\begin{tabular}{lcc}
	\toprule
	 & \makecell[c]{No extra \\parameters} & \makecell[c]{Ensembled \\knowledge} \\
    \midrule
    Self-distillation \cite{furlanello2018born,zhang2019your,xu2019data,yun2020regularizing,kim2021self} & \textcolor{ForestGreen}{\ding{52}} & \textcolor{red}{\ding{56}} \\
    Ensemble distillation \cite{lan2018one,shen2019meal,tian2019crd,guo2020online,son2021densely} & \textcolor{red}{\ding{56}} & \textcolor{ForestGreen}{\ding{52}} \\
    Label refinery \cite{bagherinezhad2018label,beyer2020we,yun2021relabel} & \textcolor{red}{\ding{56}} & \textcolor{red}{\ding{56}} \\
    Our \textbf{BAKE} & \textcolor{ForestGreen}{\ding{52}} & \textcolor{ForestGreen}{\ding{52}} \\
    \toprule
	\end{tabular}
	\end{center}
	\vspace{-15pt}
    \caption{Key differences between our method and related works.}
    \vspace{-10pt}
	\label{tab:diff}
\end{table}

\vspace{-10pt}
\paragraph{Knowledge ensembling.}
It is well-known that an ensemble of multiple networks generally yields better predictions than a single network in the ensemble. The ensembling technologies aim to generate robust supervision signals via aggregating models \cite{meanteacher,he2020momentum,byol} or predictions \cite{temensemble,lan2018one,shen2019meal,tian2019crd,guo2020online,son2021densely}. Several attempts leveraged the spirit of knowledge ensembling in distillation tasks, dubbed ``ensemble distillation'' methods. For example, CRD \cite{tian2019crd} and MEAL \cite{shen2019meal} proposed to enhance the soft targets by ensembling the knowledge of multiple pre-trained teacher networks. KDCL \cite{guo2020online} introduced to aggregate the information from multiple independent students, which are collaboratively training.
The idea of knowledge ensembling can be not only applied in supervised tasks, but also employed in semi-supervised \cite{temensemble,meanteacher} and self-supervised \cite{he2020momentum,byol} learning tasks.
Note that the way of ``knowledge ensembling'' is not limited to multi-model ensembling, \textit{e.g.}, in semi/self-supervised learning tasks, \cite{meanteacher,he2020momentum,byol} use temporal ensembling with momentum updates to integrate the knowledge of a student model over iterations.
Despite a big success, we remark that existing knowledge ensembling techniques all require additional networks or branches, which may be inapplicable in resource-limited environments. 

\vspace{-10pt}
\paragraph{Label refinery for ImageNet classification.}
ImageNet \cite{deng2009imagenet} is a widely-acknowledged dataset in computer vision. Although it could well benchmark the performance of image classification methods, some studies \cite{bagherinezhad2018label,beyer2020we,shankar2020evaluating,yun2021relabel} have observed that the manually annotated labels for ImageNet are incomplete. Specifically, ImageNet was annotated under a single-labeling policy, \textit{i.e.}, one label per image, however, there are generally multiple objects in a single image. To overcome the problem, state-of-the-art label refinery methods \cite{bagherinezhad2018label,yun2021relabel} produced multi-labels by an auxiliary annotator. For instance, \cite{bagherinezhad2018label} introduced an iterative training scheme, 
\textit{i.e.}, the trained network acted as the annotator for the next generation training.
\cite{yun2021relabel} relabeled the dataset with a super-strong classifier, which was pre-trained on super-scale datasets. Although they cleaned up the noisy labels to some extent, they heavily depended on the capacity of external networks and required much resources to train a strong annotator, which was inflexible.

% Besides the fully-supervised ImageNet classification, label refinery for pseudo-labeling is also critical in semi-supervised learning tasks \cite{}. \cite{} introduced to generate refined pseudo labels for unlabeled data by propagating the labels from their nearest neighbors. 

% TODO LABEL PROPAGATION AND RANDOM WALK

The key advantages of our BAKE against existing self-distillation, ensemble distillation and label refinery methods are summarized in Table \ref{tab:diff}.

\begin{figure*}[t]
\centering
\includegraphics[width=0.9\linewidth]{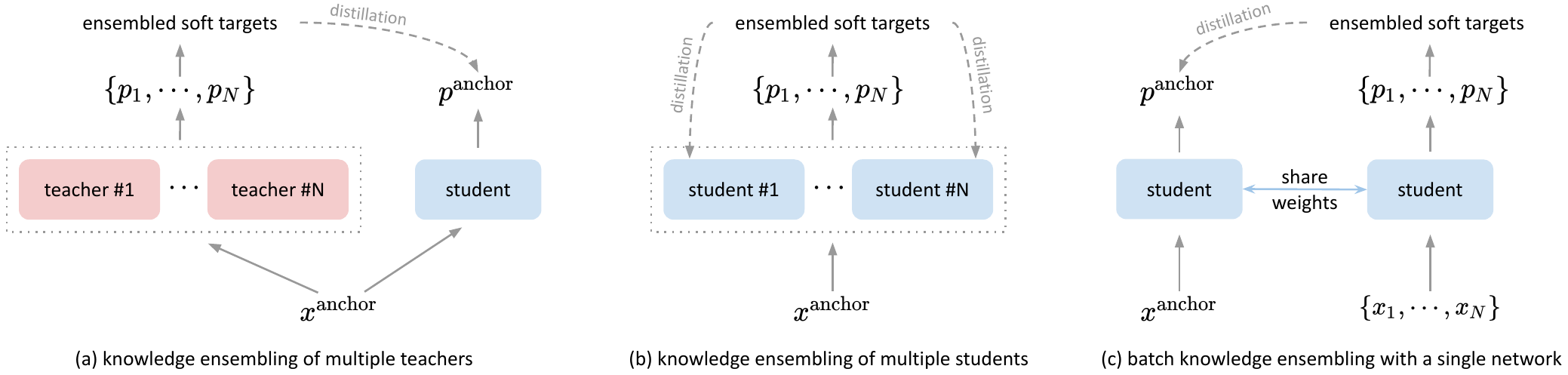}
\vspace{-5pt}
\caption{Conceptual comparison of three knowledge ensembling mechanisms. (a) {\bf Multi-teacher ensembling} \cite{shen2019meal,tian2019crd}: the soft targets are produced by ensembling the knowledge from multiple pre-trained high-capacity teachers. (b) {\bf Multi-student ensembling} \cite{guo2020online}: the soft targets are refined by ensembling the knowledge from multiple peer-teaching students. (c) {\bf The proposed BAKE}: generating robust soft targets by ensembling the knowledge of the other samples (\textit{i.e.}, $\{x_1,\cdots,x_N\}$) in the same mini-batch. BAKE enables knowledge ensembling \textit{with a single network}, saving the memory and time costs to a large extent.}
\vspace{-5pt}
\label{fig:compare}
\end{figure*}

\section{Method}
% \vspace{-2pt}
\subsection{Revisit of Knowledge Distillation}
\vspace{-2pt}

% Knowledge distillation can be thought of as training the network for a mimic task.

% Given a single network,
% with parameters $\theta$, 
Knowledge distillation \cite{kd} can be considered as regularizing the training of the student network using soft targets that carry the ``dark knowledge'' of the teacher network.
% The teacher network could be a 
% training a student network for a mimic task, \textit{i.e.}, approaching the teacher network via soft targets.

% aim to regularize the training using the teacher's ``dark knowledge'' as the soft learning targets.
% The ``teacher'' can be cast as a pre-trained strong network, a mutual-learning student, or even the distilled network itself.

The student network generally consists of a backbone encoder $F$ and a classifier $C$ to perform classification.
For each training sample $x$, its logit vector is encoded as $\vz=C(F(x))$. 
The predictive probability vector $\vp^\tau$ can be obtained via a softmax function on the logits, \ie, the probability of class $k$ can be formulated as
% \vspace{-5pt}
\begin{align}
    \vp^\tau(k)=\frac{\exp(\vz_k/\tau)}{\sum_{i=1}^K \exp(\vz_i/\tau)},
\end{align}
where $\tau$ is a temperature hyper-parameter,
% $z_k$ denotes the logit of the $k$th class 
and $K$ is the number of total classes.
Let $y\in\{1,\dots,K\}$ denotes the ground truth label and $\vq^\tau$ is the soft target produced by the teacher network.
% online generated by the current network.
The cross-entropy loss and the KL divergence between the predictions and soft targets are minimized jointly to train the student via
% \vspace{-5pt}
\begin{align}
\label{eq:self-distill}
    \mathcal{L}_x=-\log \vp(y)+\lambda\cdot\tau^2\cdot D_\text{KL}(\vq^\tau\|\vp^\tau),
\end{align}
where $\vp(y)$ denotes the probability normalized without a temperature, and $\lambda$ weights the two terms.

Recent works \cite{lan2018one,shen2019meal,tian2019crd,guo2020online} found that ensembling diverse ``dark knowledge'' from multiple teachers or students can form better soft targets, leading to better final performance (see Figure \ref{fig:compare} (a)\&(b) for details). 
However, this strategy would increase much more computational and memory overhead to enable multiple networks or branches training. 
% The soft targets $\vq^\tau$ can be produced by intermediate layers \cite{}, last-generation networks \cite{}, or intra-class samples \cite{}. 
% In general, soft targets from more accurate teachers would lead to better final performance. 
% However, INTRODUCE A PROBLEM HERE TO MOTIVATE YOUR METHOD. 
To tackle the challenge, we introduce batch knowledge ensembling in a single network via self-distillation, as illustrated in Figure \ref{fig:compare} (c).
% Note that existing self-distillation methods cannot achieve

% the batch knowledge ensembling scheme to generate more accurate soft targets by propagating knowledge between samples with a single network.

\subsection{Batch Knowledge Ensembling}
\vspace{-2pt}

% Existing works \cite{} aggregate diverse ``dark knowledge'' from multiple teachers to form better soft targets at the cost of much more computational and memory overhead. SELF-DISTILLATION WITH ONE NETWORK DOESN'T HAVE THE ISSUE?

Rather than utilizing multiple pre-trained high-capacity teachers \cite{shen2019meal,tian2019crd} or collaboratively training students \cite{guo2020online}, 
we propose a novel self-knowledge ensembling solution by exploring how to ensemble knowledge of different samples in the same mini-batch with a single network, \ie, the student network itself.
% \textit{i.e.}, the soft targets are ensembled and produced using the student's own ``dark knowledge''.
% We propose a novel solution knowledge ensembling by exploring knowledge of other samples within the same batch and without needing multiple networks.
%Rather than ensembling multiple teacher networks, we introduce to ensemble the knowledge of different samples in a mini-batch.
Intuitively, samples with high visual similarities are expected to have more consistent predictions on their predicted class probabilities, regardless of their ground-truth labels. In our solution, similar samples' knowledge is systematically aggregated and ensembled to provide better soft targets.

\vspace{-10pt}
\paragraph{Batch knowledge propagation and ensembling.}
We propose to propagate and ensemble knowledge among samples on-the-fly in terms of their feature similarities.
% for generating more accurate soft targets.
Given a mini-batch of $N$ samples and a network $C\circ F$ under training, we first estimate the samples' pairwise similarities by the dot product of their encoded representations with the current network. Such similarities can be stored in an affinity matrix $A\in\mathbb{R}^{N\times N}$ as
% \vspace{-5pt}
\begin{align}
    A(i,j)=\sigma(F(x_i))^\top \sigma(F(x_j)),
\end{align}
where $\sigma(\rvf)=\rvf/\|\rvf\|_2$ denotes the $\ell_2$-norm and ${i,j}$ are the indices of samples in a batch.
To avoid the self-knowledge reinforcement, we discard the diagonal entries from $A$ by $A=A\odot(1-I)$, where $I$ is an identity matrix and $\odot$ denotes the element-wise multiplication.
Subsequently, we normalize each row of the affinity matrix $A$ so that $\sum_{j=1}^N \hat{A}(i,j)=1$ for all $i$, while keeping the diagonal all zeros, \textit{i.e.}, $\hat{A}(i,i)=0$.
% , while discarding the diagonal entries. 
The normalization can be formulated as a softmax function over each row of the matrix $A$
% \vspace{-5pt}
\begin{align}
    \hat{A}(i,j) = \frac{\exp(A(i,j))}{\sum_{j\ne i} \exp(A(i,j))},~~\forall i\in\{1,\dots,N\}.
\end{align}
% \vspace{-5pt}
% Note that all the diagonal entries of $\hat{\mA}$ are set to zero, \textit{i.e.}, $\hat{A}(i,i)=0$, for avoiding the self-knowledge reinforcement. 
% aggregating the knowledge from other samples.
% for avoiding self-reinforcement during knowledge ensembling.
% The original $\mA$ will be replaced by $\hat{\mA}$ after the normalization, \textit{i.e.}, $\mA\leftarrow \hat{\mA}$. 

We denote the predicted probabilities of samples within a batch as $P^\tau=[\vp_1^\tau,\dots,\vp_N^\tau]^\top \in \mathbb{R}^{N\times K}$, which satisfy $\sum_{k=1}^K {P}^\tau(i,k)=1, \forall i$. 
For the $i$-th sample in the mini-batch, 
we would like to weightedly propagate and ensemble the other samples' predictions to create a better soft target for it based on the inter-sample affinities, which can be formulated as
% \vspace{-5pt}
\begin{align}
    \hat{\vp}_i^\tau=\sum_{j\ne i}\hat{A}(i,j)\vp^\tau_j=\hat{A}(i)P^\tau,
\end{align}
where $\hat{\vp}_i^\tau$ is the propagated probability vector for the $i$-th sample and can serve as the refined soft targets.
Intuitively, if the $i$-th sample and the $j$-th sample are similar with a high affinity $\hat{A}(i,j)$, the prediction $\vp^\tau_j$ would have a larger weight to be propagated to $\hat{\vp}_i^\tau$.
% For propagating the whole batch in parallel, 
Propagating the predictions between all the samples in a mini-batch in parallel can be formulated as
%we could measure the propagated probability matrix via 
$\hat{P}^\tau=\hat{A} P^\tau$.
% to achieve the 
% The knowledge carried by the probability vectors can therefore be propagated via
% \begin{align}
%     \hat{\mP}^\tau=\hat{\mA} \mP^\tau,
% \end{align}
% where $\hat{\mP}^\tau$ is the 
% propagated probability matrix after one iteration. 

To avoid propagating and ensembling noisy predictions too much, 
we produce the soft learning targets $Q^\tau$ as a weighted sum of the initial probability matrix $P^\tau$ and the propagated one $\hat{A} P^\tau$, 
\begin{align}
\label{eq:once}
    Q^\tau = \omega \hat{A} P^\tau + (1-\omega)P^\tau,
\end{align}
where $\omega\in[0,1]$ is the weighting factor and 
$Q^\tau=[\vq_1^\tau,\dots,\vq_N^\tau]^\top\in\mathbb{R}^{N\times K}$.
With the above formulations,
the knowledge of the samples within the same batch can be propagated to each other and ensembled for one iteration.
% Given the above formulations, we can achieve one iteration of the knowledge (prediction) propagation and ensembling in a mini-bath.

% The knowledge from other samples in the same batch can therefore be propagated and ensembled to the anchor sample.

%%%%%%%%%%%%%%%%%% ALGORITHM %%%%%%%%%%%%%%%%%%
\SetAlCapHSkip{0em}
\setlength{\algomargin}{0pt}
\begin{algorithm}[t!]
\scriptsize
\ttfamily
\caption{PyTorch-style pseudocode for BAKE.}
\label{alg}
\algorithmfootnote{\texttt{mm}: matrix multiplication; \texttt{eye}: identity matrix; \texttt{inv}: inverse matrix.}
% ~\\
\Comment{w: ensembling weight}
\Comment{t: temperature}
\Comment{r: loss weight}
% ~\\
\fontseries{m}\selectfont for (x, gt\_labels) in loader:\\
~~~~\Comment{features: N$\times$D, logits: N$\times$K}
~~~~f, logits = net.forward(x)\\
% ~\\
~~~~\Comment{classification loss with ground-truth labels}
~~~~loss = CrossEntropyLoss(logits, gt\_labels)\\
~\\
~~~~\Comment{produce soft targets}
~~~~f = normalize(f)\\
~~~~A = softmax(mm(f, f.t())-eye(N)*1e9) \Comment{row-wise normalization of affinity matrix with zero diagonal}
~~~~soft\_targets = mm((1-w)$\cdot$inv(eye(N)-w$\cdot$A), softmax(logits/t)) \Comment{approximate inference for propagation and ensembling}
~~~~soft\_targets = soft\_targets.detach() \Comment{no gradient}
% ~\\
~~~~\Comment{distillation loss with soft targets}
~~~~loss += KLDivLoss(log\_softmax(logits/t), soft\_targets)*t$^\texttt{2}$*r\\
% ~\\
~~~~\Comment{SGD update}
~~~~loss.backward()\\
~~~~update(net.params)\\
% ~\\
% \vspace{-10pt}
\end{algorithm}
% \vspace{-10pt}
%%%%%%%%%%%%%%%%%% ALGORITHM %%%%%%%%%%%%%%%%%%

\vspace{-10pt}
\paragraph{Approximate inference.}
The knowledge propagation and ensembling can be conducted for multiple times until convergence for fully fusing their knowledge
% \vspace{-5pt}
\begin{align}
    Q^\tau_{(t)} &= \omega \hat{A} Q^\tau_{(t-1)} + (1-\omega)P^\tau \nonumber\\
    &= (\omega \hat{A})^t P^\tau + (1-\omega) \sum_{i=0}^{t-1} (\omega \hat{A})^i P^\tau,
\end{align}
 where $t$ denotes the $t$-th propagation and ensembling iteration. 
When the number of iterations approaches infinite, we have $\lim_{t\to \infty}(\omega \hat{A})^t=0$, given that $\omega\in[0,1]$.
Also, since 
% \vspace{-5pt}
\begin{align}
    \lim_{t\to \infty}\sum_{i=0}^{t-1} (\omega\hat{A})^i=(I-\omega\hat{A})^{-1},
\end{align}
where $I$ is an identity matrix. We can obtain an approximate inference formulation for the knowledge propagation and ensembling as
% \vspace{-5pt}
\begin{align}
\label{eq:infty}
    Q^\tau_{(\infty)}=(1-\omega)(I-\omega\hat{A})^{-1}P^\tau.
\end{align}
Note that $Q^\tau_{(\infty)}$ naturally satisfies $\sum_{k=1}^K Q^\tau_{(\infty)}(i,k)=1$ for all $i$ without requiring extra normalization, which forms valid soft targets for training. The gradient is not back-propagated through the soft targets $Q^\tau_{(\infty)}$ for stable training.
% to avoid the model collapse. MODEL COLLAPSE IS A SPECIFIC WORD IN GAN. FIND ANOTHER NAME.

For each training sample, 
% THE NOTATION OF THE SAMPLE X APPEARS SO LATE. NO NEED TO DEFINE X NOW. 
we estimate its soft targets $\vq^\tau_{(\infty)} \in Q^\tau_{(\infty)}$ by ensembling the knowledge from other samples in the same batch with the approximate inference formula (Eq. (\ref{eq:infty})). The produced $\vq^\tau_{(\infty)}$ is then used as a refined learning target to supervise the self-distillation procedure with Eq. (\ref{eq:self-distill}).
The overall training procedure is detailed in Algorithm \ref{alg}.

% \paragraph{Training objective.}
% We adopt a similar loss function as Eq. (\ref{eq:self-distill}). For each sample $x$, its training objective can be formulated as
% \begin{align}
% \label{eq:loss}
%     \mathcal{L}_x=-\log p(y)+\lambda\cdot\tau^2\cdot D_\text{KL}(\vq^\tau_{(\infty)}\|\vp^\tau),
% \end{align}
% where the soft target $\vq^\tau_{(\infty)} \in \mQ^\tau_{(\infty)}$.

\subsection{Discussion}
\label{sec:discuss}
% \vspace{-5pt}

% \paragraph{Effects of soft targets produced by BAKE.}
% show some cases

\paragraph{Time and memory consumption.}
BAKE generates refined soft targets on-the-fly with \textit{minimal} computational and memory overhead compared to existing ensemble distillation methods.
% and consistently surpasses all the state-of-the-art self-distillation methods \cite{}, ensemble distillation methods \cite{} and label refinery methods \cite{}.
BAKE does not need additional network parameters on top of ordinary backbones (\eg, \cite{lan2018one}), and does not require auxiliary networks (\eg, multiple student networks for peer teaching \cite{guo2020online}). 
Only very few additional GPU memory (+2.2\% for ResNet-50) is required for prediction propagation and ensembling in addition to the conventional classification loss.
Furthermore, BAKE does not require pre-trained teachers or annotators (\eg, \cite{tian2019crd,shen2019meal}). Only little additional training time (+3.7\% for ResNet-50) is required for ensembling the knowledge within each batch and optimizing the network via the self-distillation loss.
% BAKE 
%not only improves the classification performance at the minimum cost, but also
% consistently surpasses all state-of-the-art self-distillation methods \cite{}, ensemble distillation methods \cite{}, as well as label refinery methods \cite{} (will be discussed in Section \ref{}).

\vspace{-10pt}
\paragraph{Data sampling.}
BAKE propagates and ensembles the knowledge based on inter-sample affinities. In the experiments, we found that BAKE does not work if the samples in a mini-batch are totally dissimilar to each other, \ie, showing uniformly low affinities. 
To ensure that similar pairs can always be found in the mini-batch, we introduce a per-class data sampler on top of the common random sampling mechanism. Specifically, 
for an anchor image in a randomly sampled mini-batch, we randomly select another $M$ images out of the same class.
% (we use $M=1$ in our experiments).
% given a randomly sampled mini-batch, we sample another set of samples from the dataset by randomly selecting $M$ images from the same classes.
Given the initial batch size of $\hat{N}$ and $M$ intra-class images for each anchor, we would have a total batch size of $N=\hat{N}\times(M+1)$ for training.
% Since the intra-class data sampler may decrease the batch diversity on each GPU, we use
For example, we use $\hat{N}=256$ and $M=1$ for our experiments on ImageNet, yielding a batch size of $N=512$.

\section{Experiments}

\subsection{Experimental Details}

\paragraph{Datasets.}
% \subsection{Datasets}
We study the effectiveness of BAKE mainly on 
% six image classification tasks. 
the large-scale ImageNet-1K \cite{deng2009imagenet} (ILSVRC2012), which
% , which contains over 1 million training images and 50,000 validations images out of 1,000 classes, 
is considered as one of the most important benchmarks in learning visual representations.
% used as the main benchmark in our paper.
We also evaluate BAKE on the other two conventional image classification datasets, CIFAR-100 \cite{cifar} and TinyImageNet, and three fine-grained image classification datasets, CUB-200-2011 \cite{cub}, Stanford Dogs \cite{dog} and MIT67 \cite{mit}. 
% The statistics of datasets can be found in Table \ref{tab:data}.
% The number of classes and images for each dataset can be found in Table \ref{tab:data}.
Top-1 and top-5 classification accuracies are calculated for evaluation.

% \begin{table}[htb]
% \footnotesize
% \begin{center}
% % 	\vspace{-5pt}
% 	\begin{tabular}{lccc}
%     \toprule
%     Dataset & \# classes & \# train images & \# val images \\
%     \midrule
%     ImageNet \cite{deng2009imagenet} & 1,000 & 1,281,167 & 50,000 \\
%     CIFAR-100 \cite{cifar} & 100 & 50,000 & 10,000 \\
%     TinyImageNet & 200 & 100,000 & 10,000 \\
%     CUB-200-2011 \cite{cub} & 200 & 5,994 & 5,794 \\
%     Stanford Dogs \cite{dog} & 120 & 12,000 & 8,580 \\
%     MIT67 \cite{mit} & 67 & 5,360 & 1,340 \\
%     \toprule
% 	\end{tabular}
% 	\end{center}
% 	\vspace{-5pt}
%     \caption{Statistics of the datasets used for training and evaluation.}
%     \vspace{-10pt}
% 	\label{tab:data}
% \end{table}

\vspace{-10pt}
\paragraph{Network architectures.}
% We study different architectures on various datasets.
To demonstrate that our BAKE can consistently improve various architectures on multiple datasets, 
we study the family of ResNets, including ResNet \cite{he2016deep}, ResNeSt \cite{zhang2020resnest} and ResNeXt \cite{Xie2016}, 
the lightweight networks, including MobileNet-V2 \cite{howard2017mobilenets} and EfficientNet-B0 \cite{tan2019efficientnet}, 
and the vision transformers, including Swin-T/S/B \cite{liu2021swin},
on the ImageNet benchmark.
We also evaluate ResNet \cite{he2016deep} and DenseNet \cite{huang2017densely} on the other relatively smaller scale datasets.
Note that we use the pre-activation blocks \cite{he2016identity} for CIFAR-100 and TinyImageNet datasets following \cite{yun2020regularizing}.

% \paragraph{Hyper-parameters.} 
% There are three hyper-parameters for the training objective. 
% In the loss function Eq. (\ref{eq:self-distill}), we set the distillation loss weight $\lambda$ as $1$ and the temperature $\tau$ as $4$.
% In the approximate inference function Eq. (\ref{eq:infty}), we select the ensembling weight $\omega$ from $\{0.5, 0.7\}$, \textit{i.e.}, $0.5$ for ImageNet, CIFAR-100 and CUB-200-2011, and $0.7$ for TinyImageNet, Stanford Dogs and MIT67.

\subsection{ImageNet Classification}
% \vspace{-5pt}
\paragraph{Training details.} 
There are three hyper-parameters required by the training of BAKE. 
In the loss function Eq. (\ref{eq:self-distill}), we set the distillation loss weight $\lambda$ as $1.0$ and the temperature $\tau$ as $4.0$.
In the approximate inference function Eq. (\ref{eq:infty}) of knowledge ensembling, we set the ensembling weight $\omega$ as $0.5$.
BAKE is not sensitive to these hyper-parameters, which will be discussed next.
$M$ is set to $1$ for the per-class data sampling (see Section \ref{sec:discuss}).
% All the experiments on CNNs/Transformers are trained for 100/300 epochs on 8 GPUs if not specified.
If not specified, all the experiments on CNNs and Transformers are trained on 8 GPUs for 100 and 300 epochs, respectively. More details can be found at Appendix \ref{app:training_details}.

% \paragraph{Training details.}
% We implement our BAKE on top of the open-source codebase\footnote{\url{https://github.com/facebookresearch/pycls}} for CNN architectures, and follow most of the training settings of \cite{radosavovic2020designing}.
% Specifically, we use SGD as our optimizer with a momentum of $0.9$.
% We use standard augmentation techniques including random cropping, flipping, and lighting noise. All the images are resized to $224\times 224$ for training and $256\times 256$ for validation.
% The batch size is set to $512$, \ie, $\hat{N}=256$ for the initial random sampling and $M=1$ for the per-class data sampling (see Section \ref{sec:discuss}).
% The base learning rate is set as $0.05$ for MobileNet, $0.2$ for the family of ResNets and $0.4$ for EfficientNet. The initial learning rate is calculated via $\text{lr}=\text{base\_lr} \times \text{batch\_size}/256$.
% We use cosine schedule with 5-epoch warm-up as the learning rate policy.
% All the experiments on CNNs are trained for 100 epochs on 8 GPUs if not specified.
% When implementing BAKE on Swin Transformer \cite{liu2021swin}, we follow most of their original training hyper-parameters, \textit{i.e.}, all the models are trained for 300 epochs with a batch size of 1024 ($\hat{N}=512$, $M=1$) on 8 GPUs. 
% We also adopt $M=1$ for training with Swin Transformer.

\vspace{-10pt}
\paragraph{Improvements on various architectures.}
% As illustrated in Table \ref{tab:arch}, we verify the effectiveness of BAKE on multiple network architectures. 
% We not only consider the widely-used architectures (\eg, ResNet-50), but also evaluate both the deeper/wider architectures (\eg, ResNet-152, ResNeXt-152) and the lighter architectures (\eg, MobileNet).
% BAKE consistently improves the ``vanilla'' setting (training with a cross-entropy loss) by significant margins. 
% Most importantly, BAKE boosts the performance with only negligible computational overhead, \ie, for each training iteration of ResNet-50, BAKE takes only extra $+3.7\%$ more time compared to the plain classification network.
As illustrated in Table \ref{tab:arch}\&\ref{tab:vit}, we verify the effectiveness of BAKE on multiple network architectures. 
We not only consider the widely-used architectures (\textit{e.g.}, ResNet-50), but also evaluate both the deeper/wider architectures (\textit{e.g.}, ResNet-152, ResNeXt-152) and the lighter architectures (\textit{e.g.}, MobileNet). 
We also evaluate on the state-of-the-art architecture, namely Swin Transformer \cite{liu2021swin}.
BAKE consistently improves the ``vanilla'' setting (training with a cross-entropy loss) by significant margins. 
Most importantly, BAKE boosts the performance with only negligible computational overhead, \textit{i.e.}, for each training iteration of Swin-T, BAKE takes only extra $+1.5\%$ more time compared to the plain classification network.

\begin{table}[t]
\footnotesize
\begin{center}
	\begin{tabular}{lccc}
    \toprule
	\multicolumn{1}{c}{\multirow{2}{*}{Architecture}} & \multicolumn{2}{c}{Method} & \multicolumn{1}{c}{\multirow{2}{*}{GPU Time}} \\
	\multicolumn{1}{c}{} & Vanilla & Our \textbf{BAKE} & \multicolumn{1}{c}{} \\
    \midrule
    ResNet-50 & 76.8 & 78.0 (+1.2) & +3.7\% \\
    ResNet-101 & 78.6 & 79.3 (+0.7) & +2.1\% \\
    ResNet-152 & 79.1 & 79.6 (+0.5) & +1.1\% \\
    \midrule
    ResNeSt-50 & 78.4 & 79.4 (+1.0) & +2.8\% \\
    ResNeSt-101 & 79.6 & 80.4 (+0.8) & +1.3\% \\
    \midrule
    ResNeXt-101 (32x4d) & 78.7 & 79.3 (+0.6) & +1.9\% \\
    ResNeXt-152 (32x4d) & 79.3 & 79.7 (+0.4) & +1.7\% \\
    \midrule
    MobileNet-V2 & 71.3 & 72.0 (+0.7) & +1.5\% \\
    EfficientNet-B0 & 75.1 & 76.2 (+1.1) & +6.8\% \\
    \toprule
	\end{tabular}
	\end{center}
	\vspace{-15pt}
    \caption{BAKE improves various architectures with minimal computational overhead. We report the top-1 accuracy (\%) on ImageNet. ``Vanilla'' indicates training with a conventional cross-entropy loss. The time consumption is counted on 8 Titan X GPUs.}
    \vspace{-5pt}
	\label{tab:arch}
\end{table}

\begin{table}[t]
\footnotesize
\begin{center}
	\begin{tabular}{lcccc}
    \toprule
	\multicolumn{1}{c}{\multirow{2}{*}{Architecture}} & \multicolumn{1}{c}{\multirow{2}{*}{Input Size}} & \multicolumn{2}{c}{Method} & \multicolumn{1}{c}{\multirow{2}{*}{GPU Time}} \\
	\multicolumn{1}{c}{} & \multicolumn{1}{c}{} & Vanilla & Our \textbf{BAKE} & \multicolumn{1}{c}{} \\
    \midrule
    Swin-T & 224$\times$224 & 81.3 & 82.0 (+0.7) & +1.5\% \\
    Swin-S & 224$\times$224 & 83.0 & 83.3 (+0.3) & +1.0\% \\
    Swin-B & 224$\times$224 & 83.5 & 83.9 (+0.4) & +0.5\% \\
    Swin-B & 384$\times$384 & 84.5 & 84.9 (+0.4) & +0.2\% \\
    \toprule
	\end{tabular}
	\end{center}
	\vspace{-15pt}
    \caption{BAKE improves vision transformers with various scales in terms of the top-1 accuracy (\%) on ImageNet. The time consumption is counted on 8 V100 GPUs.}
    \vspace{-5pt}
	\label{tab:vit}
\end{table}

\vspace{-10pt}
\paragraph{Comparison with label regularization methods.}
As mentioned by \cite{yuan2020revisiting}, label smoothing regularization is an ad hoc distillation with manually designed soft targets. 
To indicate that the ensembled soft targets by BAKE are better roles than the ad hoc regularizations, we evaluate both conventional label smoothing \cite{szegedy2016rethinking} regularization and the newly proposed teacher-free $\text{Tf-KD}_{reg}$ \cite{yuan2020revisiting} regularization based on the ResNet-50 classification backbone network, respectively.
As shown in Table \ref{tab:self}, the above mentioned label regularization methods can improve the ``vanilla'' network with satisfactory margins, but still show inferior performance than our proposed BAKE. 
For instance, the state-of-the-art $\text{Tf-KD}_{reg}$ improves the ``vanilla'' by $0.7\%$ in terms of top-1 accuracy but is still $-0.5\%$ lower than our BAKE.

\begin{table}[t]
\footnotesize
\begin{center}
	\begin{tabular}{lccc}
	\toprule
	\multicolumn{2}{c}{\multirow{2}{*}{Method}} & \multicolumn{2}{c}{ImageNet} \\ 
	\multicolumn{2}{c}{} & top-1 acc. & top-5 acc. \\
    \midrule
    \multicolumn{2}{c}{Vanilla ResNet-50} & 76.8 & 93.4 \\
    \midrule
    Label smoothing \cite{szegedy2016rethinking} & CVPR'16 & 77.2 & 93.7 \\
    $\text{Tf-KD}_{reg}$ \cite{yuan2020revisiting} & CVPR'20 & 77.5 & 93.7 \\
    % \midrule
    BAN \cite{furlanello2018born}* & ICML'18 & 77.4 & 93.5 \\
    % \hline
    CS-KD \cite{yun2020regularizing} & CVPR'20 & 77.0 & 93.4 \\
    $\text{Tf-KD}_{self}$ \cite{yuan2020revisiting} & CVPR'20 & 77.5 & 93.7 \\
    PS-KD \cite{kim2021self} & ICCV'21 & 77.2 & 93.5 \\
    % \midrule
    % KDCL \cite{guo2020online}* & CVPR'20 & 77.8 & - \\
    % \midrule
    % ReLabel \cite{yun2021relabel} & CVPR'21 & 77.3 & 93.5 \\
    \midrule
    \multicolumn{2}{c}{Our \textbf{BAKE}} & $\bm{78.0}$ & $\bm{94.0}$ \\
    % \toprule
    % % \multicolumn{2}{c}{Vanilla ResNet-152} & 79.3 &  \\
    % % \midrule
    % PS-KD \cite{kim2021self} & ICCV'21 & 78.6 & 94.2 \\
    % % \midrule
    % \multicolumn{2}{c}{Our \textbf{BAKE}} & $\bm{79.6}$ & $\bm{94.8}$ \\
    \toprule
	\end{tabular}
	\end{center}
	\vspace{-15pt}
    \caption{Comparison with state-of-the-art label regularization and self-distillation methods, both of which are based on single networks. We report the results of ResNet-50 on the ImageNet. All the methods are reproduced on our implementation with identical training settings as BAKE for fair comparisons and their performances surpass the reported results in their original papers. (*) Note that the original BAN \cite{furlanello2018born} performs model ensembling during inference. However, it is unfair for other compared methods which only use a single model for testing. So we discard the test-time model ensembling here.}
    \vspace{-5pt}
	\label{tab:self}
\end{table}

\vspace{-10pt}
\paragraph{Comparison with self-distillation methods.}
Our BAKE can be categorized into self-distillation methods, \ie, regularizing the network predictions using its own knowledge.
State-of-the-art self-distillation methods only considered the knowledge of individual samples \cite{furlanello2018born,yuan2020revisiting,kim2021self} or pairwise samples \cite{yun2020regularizing}, producing inferior soft targets than our BAKE. To verify it, we reproduce these methods on top of our implementation and achieve better performances than the performances reported in their original papers (see Table \ref{tab:self}).
There remains an obvious performance gap between their methods and our BAKE, \eg, BAKE surpasses 
% CS-KD \cite{yun2020regularizing} by $1.0\%$ 
$\text{Tf-KD}_{self}$ \cite{yuan2020revisiting} by $0.5\%$ 
and PS-KD \cite{kim2021self} by $0.8\%$
in terms of top-1 accuracy.

\vspace{-10pt}
\paragraph{Comparison with ensemble distillation methods.}
Our BAKE is designed following the spirit of knowledge ensembling, which aims to refine the soft targets by aggregating diverse and complementary knowledge.
State-of-the-art ensemble distillation methods achieved knowledge ensembling by leveraging multiple teachers \cite{shen2019meal,tian2019crd} or multiple students \cite{guo2020online}, which are illustrated in Figure \ref{fig:compare}.
% auxiliary networks or branches. 
We argue that our BAKE could not only save the memory and time consumption by enabling knowledge ensembling in a single network, but also generate equally robust soft targets by aggregating knowledge from a set of samples in the same mini-batch.
To demonstrate it, we compare state-of-the-art MEAL \cite{shen2019meal} and KDCL \cite{guo2020online} in Table \ref{tab:ensemble}.
% by training BAKE for 200 epochs to align with the settings of KDCL. 
We observe that our BAKE achieves similar results by using a single network for much fewer training epochs.
% As illustrated in Table \ref{tab:ensemble}, BAKE outperforms KDCL by a significant gain of $+xx\%$ without using any extra networks.
BAKE even beats KDCL with half of the training epochs, where KDCL requires an extra ResNet-18 for ensembling.

\begin{table}[t]
\footnotesize
\begin{center}
	\begin{tabular}{lccc}
	\toprule
	Method & Epochs & Top-1 acc. & Extra Params \\
    \midrule
    MEAL \cite{shen2019meal} & 180 & 78.2 & ResNet-101 \& 152 \\
    KDCL \cite{guo2020online} & 200 & 77.8 & ResNet-18 \\
    Our \textbf{BAKE} & 100 & 78.0 & {None} \\
    % Our \textbf{BAKE} & 200 &  & / \\
    \toprule
	\end{tabular}
	\end{center}
	\vspace{-15pt}
    \caption{Comparison with state-of-the-art ensemble distillation methods. We report the results of ResNet-50 on the ImageNet. The results of MEAL and KDCL are from the original papers. CRD \cite{tian2019crd} and DGKD \cite{son2021densely} did not report on ResNet-50, thus is not compared here. CRD, DGKD, MEAL and KDCL require extra networks, while BAKE does not.}
    \vspace{-5pt}
	\label{tab:ensemble}
\end{table}

\vspace{-10pt}
\paragraph{Comparison with label refinery methods.}
Recent studies on improving ImageNet classification found that the single labels are noisy as there might exist multiple objects per image.
State-of-the-art ReLabel \cite{yun2021relabel} method introduced to use a super-strong EfficientNet-L2 for re-labelling the ImageNet with multiple labels in the image spatial plane. 
However, it requires super-scale datasets ($\sim$1K times larger than the original ImageNet dataset) and much more time and memory consumption to pre-train an annotator network first, which is inflexible and even inapplicable if we do not have such resources.
Despite that direct comparison with ReLabel is actually \textit{unfair} due to its much larger training set and much deeper annotator network,
% Even so, 
we are glad to find that our BAKE can surpass the ReLabel method when training for 100 epochs by using only ImageNet dataset with the ResNet-50 backbone.
As the original ReLabel was trained for 300 epochs, we reproduce it with their official code to fairly compare with BAKE for 100 epochs. 
As shown in Table \ref{tab:relabel}, ReLabel achieves $77.3\%$ top-1 accuracy when trained for 100 epochs, showing much lower performance than our BAKE's $78.0\%$.
BAKE is also well compatible with external training tricks, \textit{e.g.}, CutMix \cite{yun2019cutmix}. 
BAKE consistently improves the baseline performance by noticeable $+1.2\%$ ($77.4\% \to 78.6\%$) when trained for 100 epochs with CutMix.
``BAKE + CutMix'' also consistently surpasses ``ReLabel + CutMix'' by a $+0.2\%$ gain for 100 training epochs. 
When trained for longer epochs (\textit{i.e.} 300 epochs), ReLabel achieves slightly better performance due to its extra knowledge from super-scale training sets and extra annotator networks.
% BAKE achieves $79.4\%$ top-1 accuracy, slightly inferior to ReLabel's $80.2\%$. 
% However, note that ReLabel's annotator network used a super-scale training set, $\sim$1K times larger than the original ImageNet dataset. 
Note that our BAKE can obtain performance gains in a much more efficient and lightweight manner.

\begin{table}[t]
\footnotesize
\begin{center}
	\begin{tabular}{lccc}
	\toprule
	Method & Epochs & Top-1 acc. & Extra Params \\
    \midrule
    Vanilla & 100 & 76.8 & None \\
    ReLabel \cite{yun2021relabel} & 100 & 77.3 & EffNet-L2 \\
    Our \textbf{BAKE} & 100 & $\bm{78.0}$ & None \\
    \midrule
    Vanilla + CutMix \cite{yun2019cutmix} & 100 & 77.4 & None  \\
    ReLabel \cite{yun2021relabel} + CutMix \cite{yun2019cutmix} & 100 & 78.4 & EffNet-L2  \\
    Our \textbf{BAKE} + CutMix \cite{yun2019cutmix} & 100 & $\bm{78.6}$ & None  \\
    % ~ + ReLabel \cite{yun2021relabel} + AdamP & 100 & 78.4 & EffNet-L2  \\
    % ~ + Our \textbf{BAKE} + AdamP & 100 & $\bm{78.6}$ & /  \\
    \midrule
    ReLabel \cite{yun2021relabel} + CutMix \cite{yun2019cutmix} & 300 & $\bm{80.2}$ & EffNet-L2  \\
    Our \textbf{BAKE} + CutMix \cite{yun2019cutmix} & 300 & 79.4 & None  \\
    % \midrule
    % ResNet-50 & 300 & 77.5 \\
    % ~ + ReLabel \cite{yun2021relabel} & 300 & 78.9 \\
    % ~ + Our \textbf{BAKE} & 300 &  \\
    % ~ + ReLabel \cite{yun2021relabel} + CutMix \cite{yun2019cutmix} & 300 & 80.2 \\
    % ~ + Our \textbf{BAKE} + CutMix \cite{yun2019cutmix} & 300 & \\
    \toprule
	\end{tabular}
	\end{center}
	\vspace{-15pt}
    \caption{Comparison with state-of-the-art label refinery method on ImageNet. We use ResNet-50 as the backbone. The results of ReLabel for 100 epochs are reproduced with the official code.
    ReLabel \cite{yun2021relabel} requires a strong annotator network (EfficientNet-L2 \cite{tan2019efficientnet}) pre-trained on super-scale datasets (JFT-300M \cite{sun2017revisiting} or Instagram-1B \cite{mahajan2018exploring}) for re-labelling, while BAKE does not. }
    \vspace{-5pt}
	\label{tab:relabel}
\end{table}

\vspace{-10pt}
\paragraph{Transfer learning.}
Apart from performing the classification task, the models trained on ImageNet can also be well transferred to the downstream tasks by fine-tuning.
To show that the ImageNet pre-trained model by BAKE could achieve better transfer learning results, we evaluate two important downstream tasks on the COCO \cite{coco} dataset.
As demonstrated in Table \ref{tab:transfer}, we use Faster-RCNN \cite{fasterrcnn} and Mask-RCNN \cite{he2017mask} with feature pyramid network \cite{lin2017feature} as base models for object detection and instance segmentation, respectively.
% The experiments are conducted on top of the open-source codebase\footnote{\url{https://github.com/open-mmlab/mmdetection}}.
The baseline results 
% (\textit{e.g.}, 37.7\% mAP for detection) 
are achieved by fine-tuning the model pre-trained with the ``vanilla'' backbone.
% ($76.8\%$ top-1 accuracy for classification).
We observe that the model pre-trained by BAKE 
% ($78.0\%$ top-1 accuracy for classification) 
could consistently improve the baseline results by $0.5\% \sim 0.8\%$ in terms of mAP.

\begin{table}[t]
\footnotesize
\begin{center}
	\begin{tabular}{lccc}
    \toprule
	\multicolumn{1}{c}{} & Faster-RCNN & \multicolumn{2}{c}{Mask-RCNN} \\
% 	\cmidrule{3-4}
	\multicolumn{1}{c}{} & bbox mAP & bbox mAP & mask mAP \\
    \midrule
    ResNet-50 & 37.7 & 38.5 & 35.0 \\
    ~ + Our \textbf{BAKE} & 38.3 (+0.6) & 39.2 (+0.7) & 35.5 (+0.5) \\
    \midrule
    Swin-T ($\times 3$ scheduler) & 45.1 & 46.0 & 41.6 \\
    ~ + Our \textbf{BAKE} & 45.9 (+0.8) &46.5 (+0.5) & 42.0 (+0.4) \\
    \midrule
    Swin-S ($\times 3$ scheduler) & 48.1 & 48.5 & 43.3 \\
    ~ + Our \textbf{BAKE} & 48.7 (+0.6) &	48.9 (+0.4)	&43.5 (+0.2) \\
    \toprule
	\end{tabular}
	\end{center}
	\vspace{-15pt}
    \caption{Transfer learning performances for object detection and instance segmentation on the COCO dataset \cite{coco}.}
    \vspace{-5pt}
	\label{tab:transfer}
\end{table}

\begin{figure}[t]
\centering
\includegraphics[width=0.65\linewidth]{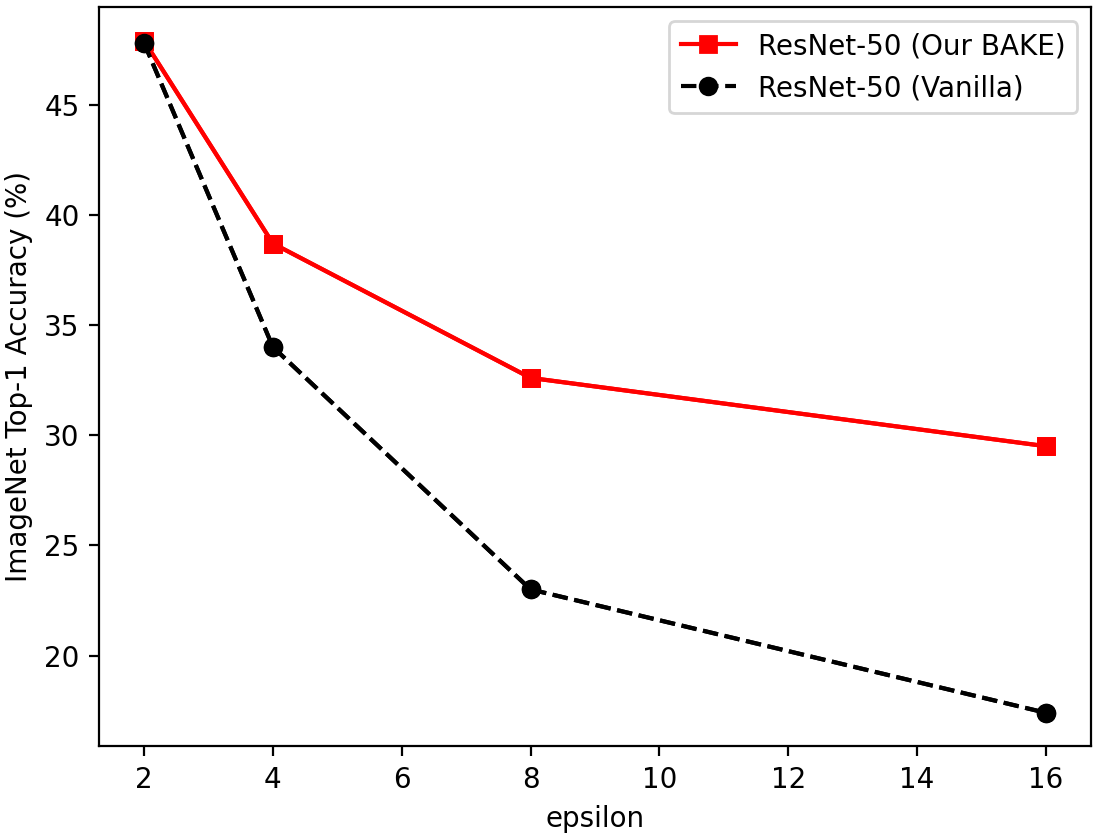}
\vspace{-10pt}
\caption{Our BAKE improves adversarial robustness against the FGSM \cite{fgsm} attack. The improvements are more significant as $\epsilon$ increases. The results are reported based on ResNet-50.}
\vspace{-5pt}
\label{fig:fgsm}
\end{figure}

\vspace{-10pt}
\paragraph{Robustness testing.}
Our BAKE does not only improve image classification, but also improves the classification robustness on much harder test sets with either common perturbations or adversarial perturbations.
ImageNet-A \cite{hendrycks2021nae} contains difficult testing images sampled from the failure cases of modern classifiers.
ImageNet-C \cite{hendrycks2019robustness} consists of 19 different corruptions and perturbations, \eg, blurring, fogging.
AutoAttack \cite{croce2020reliable} ensembles diverse parameter-free attacks.
As indicated in Table \ref{tab:robust}, BAKE successfully improves the robustness of the trained models against various perturbations.
FGSM \cite{fgsm} imposes one-step adversarial perturbations on the input image with a weight of $\epsilon$. 
As shown in Figure \ref{fig:fgsm}, at $\epsilon=16$, BAKE significantly improves ResNet-50's accuracy from $17.4\%$ to $29.5\%$ though the model is not optimized for adversarial robustness.

% BAKE improves adversarial robustness against an FGSM \cite{fgsm} attack though the model is not optimized for adversarial robustness.
\begin{table*}[t]
\scriptsize
\begin{center}
% \vspace{-10pt}
	\begin{tabular}{clcccccc}
	\toprule
	Architecture & \multicolumn{2}{c}{Method} & CIFAR-100 \cite{cifar} & TinyImageNet & CUB-200-2011 \cite{cub} & Stanford Dogs \cite{dog} & MIT67 \cite{mit} \\
    \midrule
    \multirow{6}{*}{ResNet-18} & \multicolumn{2}{c}{Vanilla} & $24.71_{\pm0.24}$ & $43.53_{\pm0.19}$ & $46.00_{\pm1.43}$ & $36.29_{\pm0.32}$ & $44.75_{\pm0.80}$\\
    % & \multicolumn{2}{c}{Vanilla (w/ sampler)} & $23.18_{\pm0.05}$ & $43.65_{\pm0.13}$ & $45.79_{\pm1.27}$ & $36.92_{\pm0.39}$ & $43.91_{\pm0.43}$\\
    \cmidrule{2-8}
    & Label smoothing \cite{szegedy2016rethinking} & CVPR'16 & $22.69_{\pm0.28}$ & $43.09_{\pm0.34}$ & $42.99_{\pm0.99}$ & $35.30_{\pm0.66}$ & $44.40_{\pm0.71}$\\
    & DDGSD \cite{xu2019data} & AAAI'19 & $23.85_{\pm1.57}$ & \bm{$41.48_{\pm0.12}$} & $41.17_{\pm1.28}$ & $31.53_{\pm0.54}$ & $41.17_{\pm2.46}$\\
    & BYOT \cite{zhang2019your} & ICCV'19 & $23.81_{\pm0.11}$ & $44.02_{\pm0.57}$ & $40.76_{\pm0.39}$ & $34.02_{\pm0.14}$ & $44.88_{\pm0.46}$\\
    & CS-KD \cite{yun2020regularizing} & CVPR'20 & $21.99_{\pm0.13}$ & $41.62_{\pm0.38}$ & $33.28_{\pm0.99}$ & $30.85_{\pm0.28}$ & $40.45_{\pm0.45}$\\
    \cmidrule{2-8}
    & \multicolumn{2}{c}{Our \textbf{BAKE}} & \bm{$21.28_{\pm0.15}$} & $41.71_{\pm0.21}$ & \bm{$29.74_{\pm0.70}$} & \bm{$30.20_{\pm0.11}$} & \bm{$39.95_{\pm0.20}$} \\
    \midrule
    \multirow{4}{*}{DenseNet-121} & \multicolumn{2}{c}{Vanilla} & $22.23_{\pm0.04}$ & $39.22_{\pm0.27}$ & $42.30_{\pm0.44}$ & $33.39_{\pm0.17}$ & $41.79_{\pm0.19}$\\
    \cmidrule{2-8}
    & Label smoothing \cite{szegedy2016rethinking} & CVPR'16 & $21.88_{\pm0.45}$ & $38.75_{\pm0.18}$ & $40.63_{\pm0.24}$ & $31.39_{\pm0.46}$ & $42.24_{\pm1.23}$\\
    & CS-KD \cite{yun2020regularizing} & CVPR'20 & $21.69_{\pm0.49}$ & $37.96_{\pm0.09}$ & $30.83_{\pm0.39}$ & $27.81_{\pm0.13}$ & $40.02_{\pm0.91}$\\
    \cmidrule{2-8}
    & \multicolumn{2}{c}{Our \textbf{BAKE}} & \bm{$20.74_{\pm0.19}$} & \bm{$37.07_{\pm0.24}$} & \bm{$28.79_{\pm1.30}$} & \bm{$27.66_{\pm0.05}$} & \bm{$39.15_{\pm0.37}$} \\
    \toprule
	\end{tabular}
	\end{center}
	\vspace{-15pt}
    \caption{Top-1 error rates (lower is better) on multiple image classification and fine-grained classification tasks. The performances of state-of-the-art single-network methods (\textit{i.e.}, label regularization methods and self-distillation methods) are reported for comparison. All the experiments are run for three times with different random seeds.}
    \vspace{-5pt}
	\label{tab:small}
\end{table*}

\begin{table}[t]
\scriptsize
\begin{center}
	\begin{tabular}{lcccc}
    \toprule
	\multicolumn{1}{c}{} & ImageNet & ImageNet-A & ImageNet-C & AutoAttack \\
	\multicolumn{1}{c}{} & top-1 acc. $\uparrow$ & top-1 acc. $\uparrow$ & mCE $\downarrow$ & top-1 acc. $\uparrow$ \\
    \midrule
    ResNet-50 & 76.8 & 2.7 & 57.9 & 1.4 \\
    ~ + Our \textbf{BAKE} & 78.0 (+1.2) & 4.6 (+1.9) & 57.4 (-0.5) & 1.9 (+0.5) \\
    \toprule
	\end{tabular}
	\end{center}
	\vspace{-15pt}
    \caption{Robustness on ImageNet-A \cite{hendrycks2021nae}, ImageNet-C \cite{hendrycks2019robustness} and AutoAttack (Linf-norm, $\epsilon=4/255$) \cite{croce2020reliable} test sets. Note that mCE is the weighted average of top-1 error rates with different corruptions (lower is better).}
    \vspace{-5pt}
	\label{tab:robust}
\end{table}

\vspace{-10pt}
\paragraph{Ablation studies on per-class data sampling.}
As described in Section \ref{sec:discuss}, we adopt a per-class data sampling strategy to ensure that the sample affinities are not uniformly low in a mini-batch.
% We found that BAKE does not work if 
We would like to claim that 
(1) The per-class data sampling strategy is critical to the success of BAKE. BAKE would fail when a conventional random sampling scheme is used.
(2) The gains of BAKE derive from the refined soft targets by ensembled knowledge rather than the data sampling strategies. 

To demonstrate our first claim, we conduct experiments when removing the proposed per-class sampling strategy. As shown in Table \ref{tab:ins}, we use ``$M=0$'' to indicate the conventional random sampling without per-class selection, and we observe that BAKE hardly improves the baseline performances.
When adopting the introduced per-class sampling with $M=1$ or $M=3$, BAKE stably boosts the baseline results by up to $+1.8\%$ gains.
As BAKE achieves similar improvements when $M>0$, we set $M$ as $1$ for brevity.

% We claim that the gains of BAKE come from the refined soft targets by ensembled knowledge rather than the data sampling strategies. 
To further verify the second claim, we evaluate ``vanilla'' when using different values of $M$.
From Table \ref{tab:ins}, we observe that the performance of ``vanilla'' settings  would decrease when using $M>0$, indicating that the per-class sampling would even hurt the model training if not used with BAKE. 
The reason might be that the per-class data sampling actually decreases the data variations within each batch.
% Since the intra-class sampling only benefits the training of BAKE, we set $M$ as $0$ when training without BAKE, \textit{e.g.}, the ``vanilla'' experiments and some comparable experime
As the ``vanilla'' experiment achieves the optimal performance when $M=0$, we do not use the per-class sampling for all the ``vanilla'' experiments throughout the paper.

\begin{table}[t]
\scriptsize
\begin{center}
	\begin{tabular}{lcccc}
	\toprule
	Method & $M$ & $\hat{N}$ & \# Batch & ImageNet top-1 acc.  \\
    \midrule
    \textbf{ResNet-50} & \textbf{0} & \textbf{256} & \textbf{256} & \textbf{76.8} \\
    ~ + Our {BAKE} & 0 & 256 & 256 & 76.8 (+0.0)  \\
    \midrule
    ResNet-50 & 1 & 256 & 512 & 76.3 \\
    \textbf{~ + Our {BAKE}} & \textbf{1} & \textbf{256} & \textbf{512} & \textbf{78.0 (+1.7)}  \\
    \midrule
    ResNet-50 & 3 & 256 & 1024 & 76.1 \\
    ~ + Our {BAKE} & 3 & 256 & 1024 & 77.9 (+1.8)  \\
    \toprule
	\end{tabular}
	\end{center}
	\vspace{-15pt}
    \caption{Ablation studies on the value of $M$ in the per-class data sampling scheme, where $\text{batch\_size}=\hat{N}\times(M+1)$.}
    \vspace{-5pt}
	\label{tab:ins}
\end{table}

\begin{table}[t]
\scriptsize
\begin{center}
	\begin{tabular}{ccc}
	\toprule
	Soft targets & $\omega$ & ImageNet top-1 acc.  \\
    \midrule
    Equal to raw predictions & 0.0 & 76.3 \\
    Refined by BAKE & 0.1 & 77.7 \\
    Refined by BAKE & 0.3 & 77.8 \\
    Refined by BAKE & 0.5 & \bm{$78.0$} \\
    Refined by BAKE & 0.7 & 77.8 \\
    Refined by BAKE & 0.9 & \bm{$78.0$} \\
    Refined by BAKE & 1.0 & 77.9 \\
    \toprule
	\end{tabular}
	\end{center}
	\vspace{-15pt}
    \caption{Ablation studies on the value of the ensembling weight $\omega$ in Eq. (\ref{eq:infty}). We report the results of ResNet-50 on the ImageNet.}
    \vspace{-5pt}
	\label{tab:ensemble_weights}
\end{table}

\vspace{-10pt}
\paragraph{Ablation studies on the ensembling weight $\omega$.}
The ensembling weight $\omega\in[0,1]$ is adopted to weigh the original knowledge of the anchor sample and the propagated knowledge of other samples within the same batch, as formulated in Eq. (\ref{eq:infty}).
The refined soft targets $Q^\tau_{(\infty)}$ ensemble more knowledge from the other samples as $\omega$ gets larger.
As shown in Table \ref{tab:ensemble_weights},
when $\omega$ approaches $0$, the soft targets is equal to the vanilla predictions of the anchor sample, \textit{i.e.}, $Q^\tau_{(\infty)}=P^\tau$, the distillation loss then becomes useless.
When $\omega$ approaches $1$, the soft targets are totally produced with propagated knowledge from the other samples in the same mini-batch. To avoid the soft targets becoming all zeros after propagating infinite iterations, we adopt only one iteration for $\omega=1$.
% become all zeros and BAKE fails.
The performance of BAKE is robust when changing the value of $\omega$ within a large interval of $[0.1,1.0]$.

% \begin{table}[t]
% \footnotesize
% \begin{center}
% 	\begin{tabular}{cc}
% 	\toprule
% 	Loss weight $\lambda$ & Top-1 acc.  \\
%     \midrule
%     % 0.1 & \\
%     0.5 & 77.8 \\
%     1.0 & \bm{$78.0$} \\
%     2.0 & \bm{$78.0$} \\
%     3.0 & \bm{$78.0$} \\
%     4.0 & 77.7 \\
%     \toprule
% 	\end{tabular}
% 	\end{center}
% 	\vspace{-5pt}
%     \caption{Ablation studies on the value of the weighting factor $\lambda$ in Eq. (\ref{eq:self-distill}). We report the results of ResNet-50 backbone on ImageNet.}
%     \vspace{-5pt}
% 	\label{tab:loss_weight}
% \end{table}

\subsection{Small-scale Dataset Classification}

% \paragraph{Hyper-parameters.} 
% There are three hyper-parameters required by the training of BAKE. 
% In the loss function Eq. (\ref{eq:self-distill}), we set the distillation loss weight $\lambda$ as $1$ and the temperature $\tau$ as $4$.
% In the approximate inference function Eq. (\ref{eq:infty}) of knowledge ensembling, we set the ensembling weight $\omega$ as $0.5$.
% BAKE is not sensitive to these hyper-parameters, which will be discussed next.

% \vspace{-10pt}
\paragraph{Training details.}
We adopt almost the same settings as those used for training the ImageNet.
We use $\lambda=1.0$ and $\tau=4.0$ for Eq. (\ref{eq:self-distill}).
The ensembling weight $\omega$ (Eq. (\ref{eq:infty})) is chosen from $\{0.5,0.9\}$, according to the results in Table \ref{tab:ensemble_weights}.
% We set $M$ as $3$ for CIFAR-100 and CUB-200-2011 datasets, and set $M$ as $1$ for the rest.
$M$ is chosen from $\{1,3\}$, according to the results in Table \ref{tab:ins}.
% The experiments are implemented on top of the open-source code\footnote{\url{https://github.com/alinlab/cs-kd}}.
% % to fairly compare with state-of-the-art \cite{yun2020regularizing}.
% Specifically, we use SGD as our optimizer with a momentum of $0.9$.
% % We use standard augmentation techniques including random cropping, flipping, and color jittering. All the images are resize to $224\times 224$ for training and $256\times 256$ for validation.
% The batch size is set to $128$ for CIFAR-100 and TinyImageNet, and set to $32$ for the fine-grained classification datasets. 
% The initial learning rate is set to $0.1$ and is decreased to $1/10$ of its previous value at the $100$-th and $150$-th epoch in the overall 200 training epochs.
% \textit{i.e.}, $N=256$ for the initial random sampling and $M=1$ for the intra-class sampling (see Section \ref{sec:discuss}).
% The base learning rate is set as $0.05$ for MobileNet, $0.2$ for the family of ResNets and $0.4$ for EfficientNet. The initial learning rate is calculated via $\text{lr}=\text{base\_lr} \times \text{batch\_size}/256$.
% We use cosine schedule with 5 epoch gradual warm-up as the learning rate policy.
All the experiments are trained with only 1 GPU. More details can be found at Appendix \ref{app:training_details}.
% s if no specified.

\vspace{-10pt}
\paragraph{Improvements on various architectures and datasets.}
As shown in Table \ref{tab:small}, we study the effectiveness of BAKE with a lightweight ResNet-18 and a deep DenseNet-121 on multiple classification datasets.
BAKE consistently improves the baseline results (``vanilla'') by significant margins, \eg, on the fine-grained dataset CUB-200-2011, BAKE boosts the baseline by $16.26\%$ with ResNet-18.
Also, on the widely-used CIFAR-100 dataset, the performance of ResNet-18 is improved by $3.43\%$ by BAKE.

\vspace{-10pt}
\paragraph{Comparison with self-distillation methods.}
Following the benchmark used by \cite{yun2020regularizing},
we compare with state-of-the-art self-distillation methods, DDGSD \cite{xu2019data}, BYOT \cite{zhang2019your} and CS-KD \cite{yun2020regularizing}.
BAKE stably surpasses all the methods except for the experiments of ResNet-18 on TinyImageNet, which shows a slight drop from DDGSD \cite{xu2019data}.
The positive results of BAKE are enough to demonstrate its effectiveness and superiority over existing self-distillation methods.

% \vspace{-3pt}
\section{Limitations and Conclusions}
\vspace{-5pt}

In this work, we introduce a novel batch knowledge ensembling method, dubbed BAKE, to produce refined soft targets for self-distillation. 
% Rather than aggregating multiple networks or branches, BAKE is proposed to weightedly propagate and ensemble the knowledge from different samples in the same mini-batch. 
BAKE improves the image classification performance with minimal computational and memory overhead, and outperforms state-of-the-art single-network methods on all the tested benchmarks. 
Beyond the classification tasks, the spirit of batch knowledge ensembling has great potential on other tasks, \eg, image retrieval, segmentation and detection. 

BAKE does not work if the samples in a mini-batch are totally dissimilar to each other, so we introduce a per-class data sampler to solve this problem. 
However, the data sampler would increase the CPU time when the training dataset is large-scale since it needs to reorganize the data loader before each epoch. 

% Beyond the classification tasks, the spirit of batch knowledge ensembling has great potential on other tasks, \eg, image retrieval, segmentation and detection. Furthermore, it may also be interesting to explore the batch knowledge in modern self-supervised learning frameworks (\textit{e.g.}, \cite{he2020momentum,chen2020simple,byol}), as the batch knowledge ensembling is based on the sample affinities regardless of their ground-truth labels.
% which needs to be explored.

%%%%%%%%% REFERENCES
% {\small
% \bibliographystyle{ieee_fullname}
% \bibliography{egbib}
% }

% \clearpage
% \setcounter{page}{1}
\appendix
\section{Appendix}

\subsection{Implementation Details}
\label{app:training_details}

\paragraph{Dataset statistic.}
We evaluate our BAKE on six datasets, as demonstrated in Table \ref{tab:data}. The large-scale ImageNet \cite{deng2009imagenet}, CIFAR-100 \cite{cifar} and TinyImageNet\footnote{\url{https://tiny-imagenet.herokuapp.com}} are for conventional image classification, while CUB-200-2011 \cite{cub}, Stanford Dogs \cite{dog} and MIT67 \cite{mit} focus on fine-grained image classification tasks.
% The statistics of datasets can be found in Table \ref{tab:data}.
% The number of classes and images for each dataset can be found in Table \ref{tab:data}.
% Top-1 and top-5 classification accuracies are calculated for evaluation.

\begin{table}[htb]
\footnotesize
\begin{center}
% 	\vspace{-5pt}
	\begin{tabular}{lccc}
    \toprule
    Dataset & \# classes & \# train images & \# val images \\
    \midrule
    ImageNet \cite{deng2009imagenet} & 1,000 & 1,281,167 & 50,000 \\
    CIFAR-100 \cite{cifar} & 100 & 50,000 & 10,000 \\
    TinyImageNet & 200 & 100,000 & 10,000 \\
    CUB-200-2011 \cite{cub} & 200 & 5,994 & 5,794 \\
    Stanford Dogs \cite{dog} & 120 & 12,000 & 8,580 \\
    MIT67 \cite{mit} & 67 & 5,360 & 1,340 \\
    \toprule
	\end{tabular}
	\end{center}
	\vspace{-15pt}
    \caption{Statistics of the datasets used for training and evaluation.}
    % \vspace{-10pt}
	\label{tab:data}
\end{table}

\paragraph{Training details on ImageNet.}
% There are three hyper-parameters required by the training of BAKE. 
% In the loss function Eq. (\ref{eq:self-distill}), we set the distillation loss weight $\lambda$ as $1.0$ and the temperature $\tau$ as $4.0$.
% In the approximate inference function Eq. (\ref{eq:infty}) of knowledge ensembling, we set the ensembling weight $\omega$ as $0.5$.
% BAKE is not sensitive to these hyper-parameters, which will be discussed next.
% \paragraph{Training details.}
We implement our BAKE on top of the open-source codebase\footnote{\url{https://github.com/facebookresearch/pycls}} for CNN architectures, and follow most of the training settings of \cite{radosavovic2020designing}.
Specifically, we use SGD as our optimizer with a momentum of $0.9$.
We use standard augmentation techniques including random cropping, flipping, and lighting noise. All the images are resized to $224\times 224$ for training and $256\times 256$ for validation.
The batch size is set to $512$, \ie, $\hat{N}=256$ for the initial random sampling and $M=1$ for the per-class data sampling.
% (see Section \ref{sec:discuss}).
The base learning rate is set as $0.05$ for MobileNet, $0.2$ for the family of ResNets and $0.4$ for EfficientNet. The initial learning rate is calculated via $\text{lr}=\text{base\_lr} \times \text{batch\_size}/256$.
We use cosine schedule with 5-epoch warm-up as the learning rate policy.
All the experiments on CNNs are trained for 100 epochs on 8 GPUs if not specified.
When integrating BAKE into Swin Transformer \cite{liu2021swin}, we follow their original training protocols\footnote{\url{https://github.com/microsoft/Swin-Transformer}} but use our proposed batch formulation, \textit{i.e.}, all the models are trained for 300 epochs with a batch size of 1024 ($\hat{N}=512$, $M=1$) on 8 GPUs. 

\paragraph{Training details on small-scale datasets.}
The experiments are implemented on top of the open-source code\footnote{\url{https://github.com/alinlab/cs-kd}}.
% to fairly compare with state-of-the-art \cite{yun2020regularizing}.
Specifically, we use SGD as our optimizer with a momentum of $0.9$.
% We use standard augmentation techniques including random cropping, flipping, and color jittering. All the images are resize to $224\times 224$ for training and $256\times 256$ for validation.
The batch size is set to $128$ for CIFAR-100 and TinyImageNet, and set to $32$ for the fine-grained classification datasets. 
The initial learning rate is set to $0.1$ and is decreased to $1/10$ of its previous value at the $100$-th and $150$-th epoch in the overall 200 training epochs.

% \subsection{Dataset Statistics}
% \label{sec:app_data}

% We evaluate our BAKE on six datasets, as demonstrated in Table \ref{tab:data}. The large-scale ImageNet \cite{deng2009imagenet}, CIFAR-100 \cite{cifar} and TinyImageNet\footnote{\url{https://tiny-imagenet.herokuapp.com}} are for conventional image classification, while CUB-200-2011 \cite{cub}, Stanford Dogs \cite{dog} and MIT67 \cite{mit} focus on fine-grained image classification.

% \begin{table}[htb]
% \footnotesize
% \begin{center}
% 	\vspace{-5pt}
% 	\begin{tabular}{lccc}
%     \toprule
%     Dataset & \# classes & \# train images & \# val images \\
%     \midrule
%     ImageNet \cite{deng2009imagenet} & 1,000 & 1,281,167 & 50,000 \\
%     CIFAR-100 \cite{cifar} & 100 & 50,000 & 10,000 \\
%     TinyImageNet & 200 & 100,000 & 10,000 \\
%     CUB-200-2011 \cite{cub} & 200 & 5,994 & 5,794 \\
%     Stanford Dogs \cite{dog} & 120 & 12,000 & 8,580 \\
%     MIT67 \cite{mit} & 67 & 5,360 & 1,340 \\
%     \toprule
% 	\end{tabular}
% 	\end{center}
% 	\vspace{-5pt}
%     \caption{Statistics of the datasets used for training and evaluation.}
%     % \vspace{-10pt}
% 	\label{tab:data}
% \end{table}

\subsection{Additional Ablation Studies}

\begin{table}[t]
\footnotesize
\begin{center}
	\begin{tabular}{ccc}
	\toprule
	Soft targets & $\tau$ & Top-1 acc.  \\
    \midrule
    Original & 1.0 & 77.9 \\
    Smoothed by 2$\times$ & 2.0 & \bm{$78.0$} \\
    Smoothed by 4$\times$ & 4.0 & \bm{$78.0$} \\
    Smoothed by 8$\times$ & 8.0 & 77.8 \\
    \toprule
	\end{tabular}
	\quad
	\begin{tabular}{cc}
	\toprule
	Weight $\lambda$ & Top-1 acc.  \\
    \midrule
    % 0.1 & \\
    0.5 & 77.8 \\
    1.0 & \bm{$78.0$} \\
    2.0 & \bm{$78.0$} \\
    3.0 & \bm{$78.0$} \\
    4.0 & 77.7 \\
    \toprule
	\end{tabular}
	\end{center}
	\vspace{-15pt}
    \caption{Ablation studies on the value of the temperature $\tau$ and the value of the weighting factor $\lambda$ in Eq. (\ref{eq:self-distill}). We report the results of ResNet-50 on the ImageNet.}
    % \vspace{-5pt}
	\label{tab:temperature}
\end{table}

% \vspace{-10pt}
\paragraph{Ablation studies on the temperature $\tau$.}
The temperature $\tau$ is adopted to scale the predicted logits and the soft targets in the distillation loss (Eq. (\ref{eq:self-distill})).
A higher value of $\tau$ leads to a smoother probability distribution over classes.
As illustrated in Table \ref{tab:temperature}, we study the effects of the temperature value on BAKE by changing $\tau$ from $1.0$ to $8.0$.
We observe that BAKE is not sensitive to the temperature and achieves robust results.

% \vspace{-10pt}
\paragraph{Ablation studies on the weighting factor $\lambda$.}
$\lambda$ is adopted to balance the cross-entropy loss and the knowledge distillation term in Eq. (\ref{eq:self-distill}), which is normally set to $1.0$ for brevity. We are interested in how much it affects the final performance. As demonstrated in Table \ref{tab:temperature}, the performance is consistent when changing $\lambda$ from $1.0$ to $3.0$, and is robust in the interval of $[0.5, 4.0]$.

\paragraph{Ablation studies on the ensembled ``knowledge''.}
To properly model the knowledge carried by the samples, we use their predictions $P^\tau$ (as shown in Eq. (\ref{eq:infty})) output by the current network. To verify that the predictions are more informative than the manually annotated ground-truth labels $Y$, we conduct an experiment by replacing the soft $P^\tau$ with one-hot $Y$ in Eq. (\ref{eq:infty}). As demonstrated in Table \ref{tab:gt}, we observe $-0.4\%$ inferior to the original version of BAKE in terms of top-1 accuracy, showing the superiority of knowledge carried by the model predictions.

\begin{table}[t]
\footnotesize
\begin{center}
	\begin{tabular}{cc}
	\toprule
	Sample Knowledge & ImageNet top-1 acc.  \\
    \midrule
    Model Predictions $P^\tau$ & \bm{$78.0$} \\
    Ground-truth Labels $Y$ & 77.6 \\
    \toprule
	\end{tabular}
	\end{center}
	\vspace{-15pt}
    \caption{Ablation studies on the ensembled knowledge. We report the results of ResNet-50 on the ImageNet.}
	\label{tab:gt}
\end{table}

\begin{figure}[t]
\centering
\includegraphics[width=0.7\linewidth]{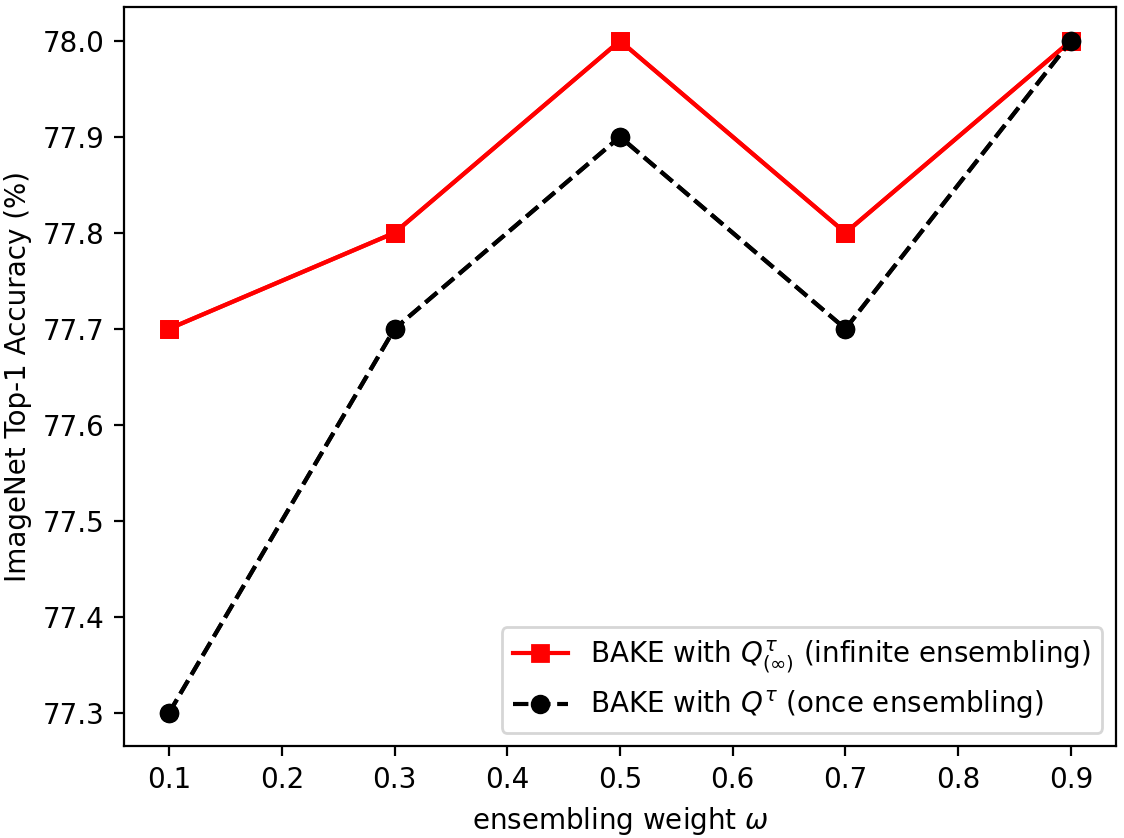}
\vspace{-5pt}
\caption{Our BAKE achieves more robust results when ensembling batch knowledge for infinite iterations. The results are reported based on ResNet-50.}
\label{fig:once}
\end{figure}

% \vspace{-5pt}
\paragraph{Ablation studies on the batch knowledge ensembling iterations.}
We adopt approximate inference to estimate the soft targets for infinite ensembling iterations. To indicate that infinite iterations achieve better learning targets than a single iteration, we conduct experiments as shown in Figure \ref{fig:once}. Specifically, ``BAKE with $Q^\tau_{(\infty)}$'' produces soft targets via approximate inference for infinite iterations (Eq. (\ref{eq:infty})) and ``BAKE with $Q^\tau$'' adopts only one iteration for ensembling (Eq. (\ref{eq:once})). We can observe that ``BAKE with $Q^\tau_{(\infty)}$'' achieves more robust results than ``BAKE with $Q^\tau$'' when changing the ensembling weight $\omega$ from $0.1$ to $0.9$.

% \vspace{-10pt}
\paragraph{Ablation studies on BN for distributed training.}
We adopt the per-class data sampler when training BAKE, as introduced in Section \ref{sec:discuss}. We find that the usage of BN is critical to final performance when employing distributed training, since the per-class data sampler decreases the data variations within each mini-batch on a single GPU. As demonstrated in Table \ref{tab:bn}, sync BN and shuffling BN \cite{he2020momentum} both work for the per-class data sampler, while using normal BN achieves even worse performance than the baseline result ($76.8\%$). We choose sync BN in our paper due to its more efficiency. Using normal BN or sync BN achieves similar performance ($\pm 0.1\%$) when training the baseline without per-class data sampling, indicating that the gains of BAKE comes from the knowledge ensembling rather than the sync/shuffling BN. Note that experiments using the per-class data sampling throughout the paper adopt sync BN during training.

\begin{table}[t]
\footnotesize
\begin{center}
	\begin{tabular}{ccc}
	\toprule
	Method & Type of BN & ImageNet top-1 acc.  \\
    \midrule
    \multirow{2}{*}{Vanilla} & Normal BN & 76.8 \\
    & Sync BN & 76.7 \\
    \midrule
    \multirow{3}{*}{Our \textbf{BAKE}} & Normal BN & 74.7 \\
    & Sync BN & 78.0 \\
    & Shuffling BN \cite{he2020momentum} & 78.1 \\
    \toprule
	\end{tabular}
	\end{center}
	\vspace{-15pt}
    \caption{Ablation studies on the usage of BN when employing distributed training. We report the results of ResNet-50 on the ImageNet.}
    % \vspace{-5pt}
	\label{tab:bn}
\end{table}

% \begin{table}[t]
% \footnotesize
% \begin{center}
% 	\begin{tabular}{ccc}
% 	\toprule
% 	Method & Epochs & ImageNet top-1 acc.  \\
%     \midrule
%     \multirow{3}{*}{Our \textbf{BAKE} + CutMix \cite{yun2019cutmix}} & 100 & 78.6 \\
%     & 200 & 79.0 \\
%     & 300 & 79.4 \\
%     \toprule
% 	\end{tabular}
% 	\end{center}
% 	\vspace{-5pt}
%     \caption{Results of BAKE for different training epochs. We report the results of ResNet-50 on the ImageNet.}
%     \vspace{-5pt}
% 	\label{tab:epoch}
% \end{table}

% \vspace{-10pt}
% \paragraph{Different training epochs of BAKE.}
% To investigate the performance of BAKE when training for different epochs, we conduct comparison experiments as demonstrated in Table \ref{tab:epoch}. All the experimental settings except the training epochs are kept the same. We observe that the results of ``BAKE + CutMix'' can be consistently boosted by training with more epochs.

\subsection{Additional Discussions}

% \paragraph{Compare with label propagation.}
% \cite{zhu2003semi,iscen2019label} introduced to propagate pseudo labels for semi-supervised learning. They assign {hard} pseudo labels by propagating the {whole dataset offline} while BAKE refines only {batches' soft targets online}, showing significant differences. What's more important, their methods are specifically designed for semi-supervised learning and are inapplicable for supervised learning.

% They are quite different. [R1-R3] are for semi-supervised learning and cannot be used for supervised learning. They assign \textit{hard} pseudo labels by propagating the {\it whole dataset offline} while BAKE refines only \emph{batches' soft targets online}. 
%$Existing label propagation cannot be used in fully supervised learning.
%showing higher efficiency especially in large-scale datasets. 
%We further introduce a per-class data sampler, that acts as a premise to the success of mini-batch propagation. 
% Plus, [43] clarified that knowledge distillation (KD) is actually a learned label smoothing (LS) regularization. BAKE produces soft targets by the online training model, which is more like a KD rather than LS.

\paragraph{Compare with random walk.}
\cite{zhou2003learning,bertasius2017convolutional} propose to use random walk to aggregate predictions, where the technique is similar to our batch knowledge ensembling. However, they use the original {hard} labels for supervision, still facing the problem of some incorrect one-hot labels.
% use random walk to aggregate predictions but use the original \emph{hard} labels for supervision, still facing the problem of some incorrect one-hot labels (see L53-92).
% They can only encourage higher prediction similarities between pixels with higher feature similarities.
%as a plug-in module to propagate the predictions, then supervised by ground-truth labels. 
In contrast, BAKE aggregates predictions to create {soft} targets for better distillation.
%In contrast, the propagated predictions serve as additional soft training targets in BAKE. 
We tested the random walk method on CIFAR-100, CUB-200-2011, as shown in Table \ref{tab:rw},
% and achieved 24.25\%, 42.51\% top-1 err. respectively, 
showing much worse performance than BAKE.
% 's 21.28\%, 29.74\%.

\begin{table}[t]
\footnotesize
\begin{center}
	\begin{tabular}{ccc}
	\toprule
	 & CIFAR-100 & CUB-200-2011  \\
    \midrule
    Vanilla & 24.71 & 46.00 \\
    Random Walk & 24.25 & 42.51 \\
    Our \textbf{BAKE} & \textbf{21.28} & \textbf{29.74}\\
    \toprule
	\end{tabular}
	\end{center}
	\vspace{-15pt}
    \caption{Comparison with random walk. The results of ResNet-18 are reported in terms of top-1 error rates (lower is better).}
	\label{tab:rw}
\end{table}

% \paragraph{Limitations.}
% BAKE does not work if the samples in a mini-batch are totally dissimilar to each other, so we introduce a per-class data sampler to solve this problem. 
% However, the data sampler would increase the CPU time when the training dataset is large-scale since it needs to reorganize the data loader before each epoch. 

\subsection{Visualization}

\begin{figure*}[t]
    \centering
    \footnotesize
    % \captionsetup{type=figure}
    \begin{tabular}{c|c|c}
        \includegraphics[width=0.35\linewidth]{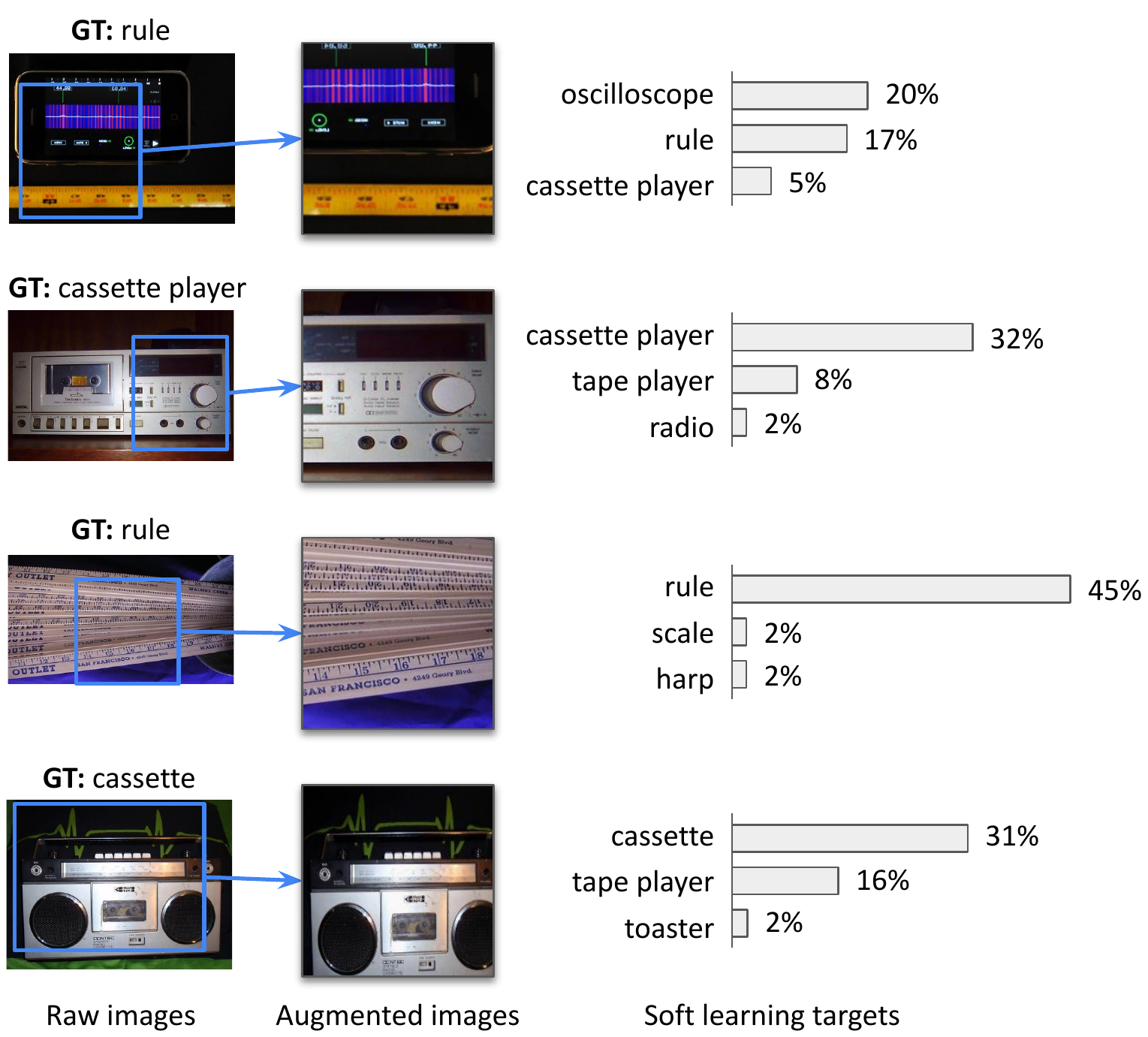}  &
        \includegraphics[width=0.3\linewidth]{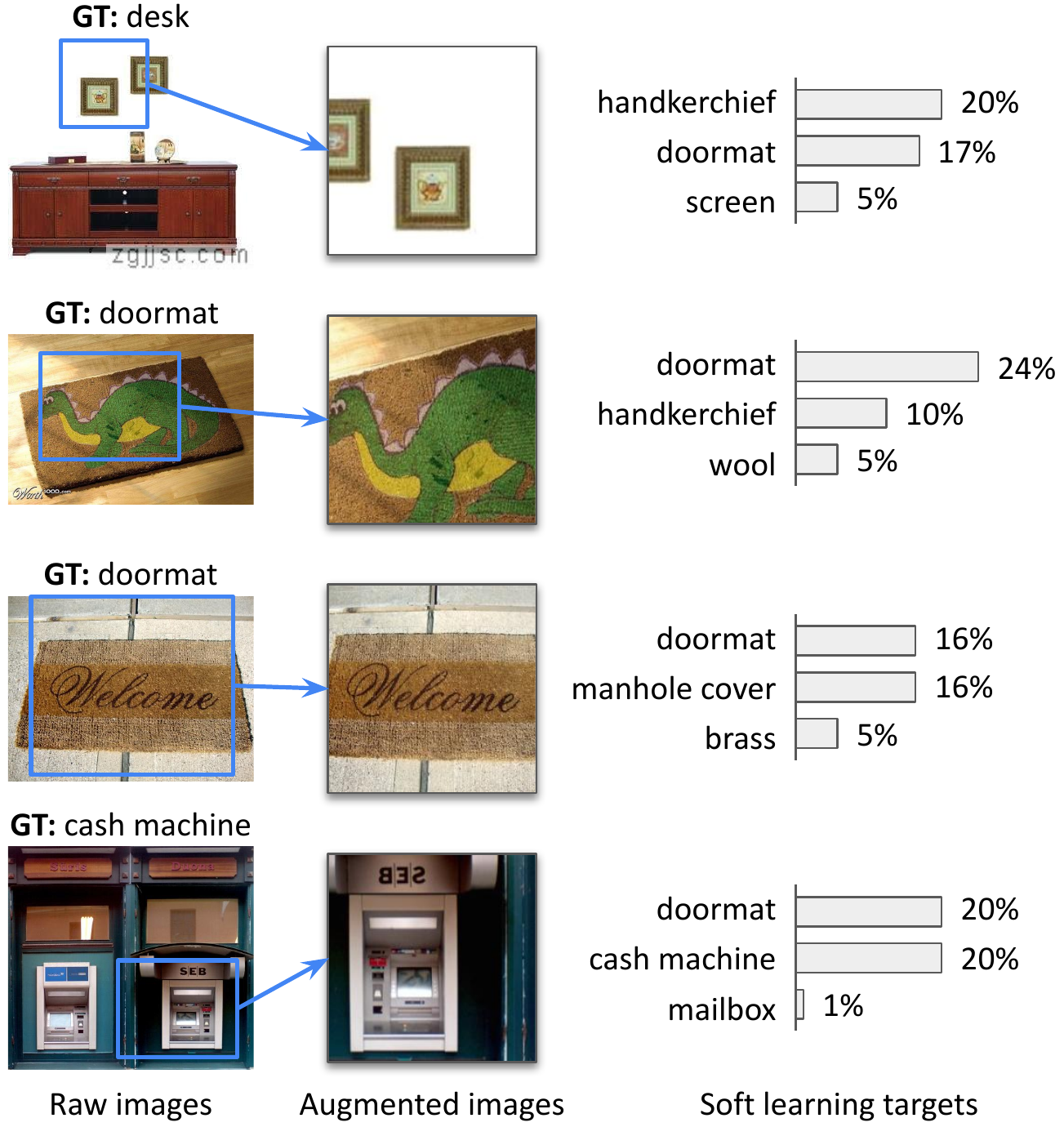} &
        \includegraphics[width=0.33\linewidth]{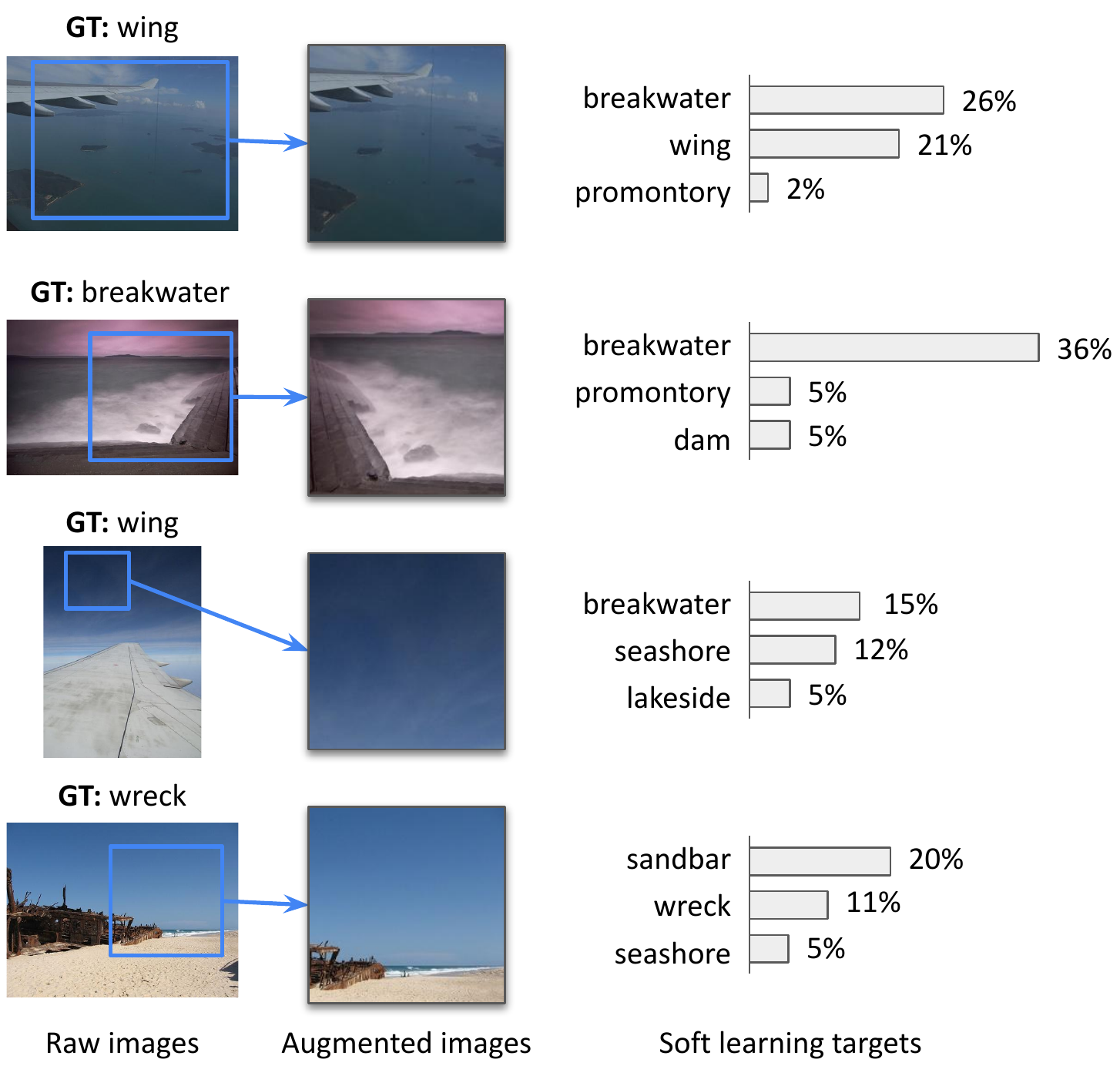} \\
        (a) Example \#1 of soft targets by BAKE & (b) Example \#2 of soft targets by BAKE & (c) Example \#3 of soft targets by BAKE \\
    \end{tabular}
    % \vspace{2pt}
    \caption{We sample three tuples of images (four images in each tuple) from three batches to show the soft targets produced by BAKE. The images are sampled from ImageNet \cite{deng2009imagenet}. ``GT'' denotes the manually annotated ground-truth labels. The knowledge of samples from the same batch is propagated and ensembled to form a better soft learning target for each sample in the batch. Note that only the top-$3$ classes of soft targets with the highest probabilities are illustrated for brevity.}
    \label{fig:softlabel}
\end{figure*}

\paragraph{Examples of soft targets.}
We illustrate the examples of soft targets produced by BAKE in Figure \ref{fig:softlabel}. 
We sample the images from three different batches, where the cross-sample knowledge propagation and ensembling are performed in each mini-batch.
There are 512 images included in each batch, and we randomly select four of them for illustration.
We present their ground-truth labels as well as the soft labels generated by BAKE. 
The soft label is a $1000$-dim probability vector ($1000$ classes for ImageNet), and we only show the probabilities of top-$3$ classes for brevity.  
It can be observed that the soft targets produced by BAKE provide more informative and complete training supervisions than the manual annotations.

\begin{figure}[h!]
    \vspace{-5pt}
    \centering
    \includegraphics[width=0.45\textwidth]{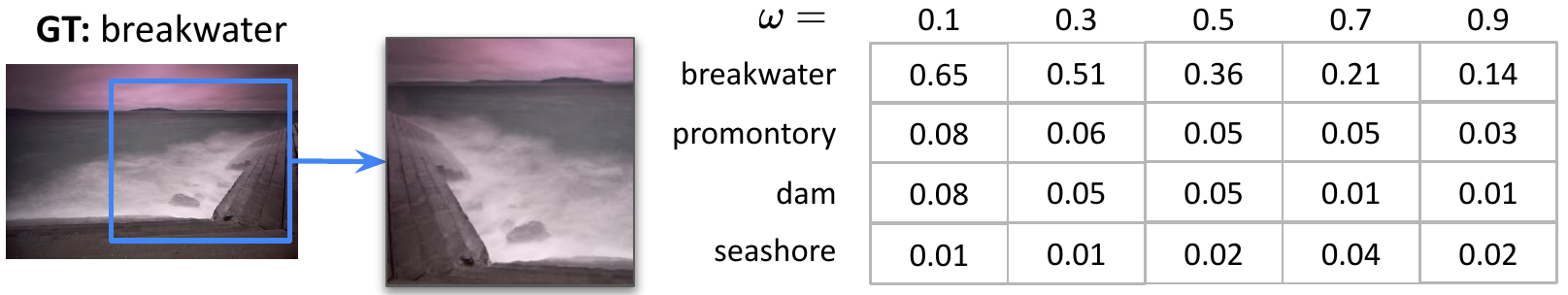}
    \vspace{-5pt}
    \caption{The illustration of soft targets with varying $\omega$. The images are sampled from ImageNet \cite{deng2009imagenet}. ``GT'' denotes the manually annotated ground-truth labels.}
    \label{fig:R1}
\end{figure}

\paragraph{Soft targets with varying $\omega$.}
% Below is a sample, and more will be added to the final version.
As illustrated in Figure \ref{fig:R1},
labels become smoother with a larger value of $\omega$.

%%%%%%%%% REFERENCES
{\small
\bibliographystyle{ieee_fullname}
\bibliography{egbib}

\begin{thebibliography}{10}\itemsep=-1pt

\bibitem{bagherinezhad2018label}
Hessam Bagherinezhad, Maxwell Horton, Mohammad Rastegari, and Ali Farhadi.
\newblock Label refinery: Improving imagenet classification through label
  progression.
\newblock {\em arXiv preprint arXiv:1805.02641}, 2018.

\bibitem{bertasius2017convolutional}
Gedas Bertasius, Lorenzo Torresani, Stella~X Yu, and Jianbo Shi.
\newblock Convolutional random walk networks for semantic image segmentation.
\newblock In {\em CVPR}, pages 858--866, 2017.

\bibitem{beyer2020we}
Lucas Beyer, Olivier~J H{\'e}naff, Alexander Kolesnikov, Xiaohua Zhai, and
  A{\"a}ron van~den Oord.
\newblock Are we done with imagenet?
\newblock {\em arXiv preprint arXiv:2006.07159}, 2020.

\bibitem{croce2020reliable}
Francesco Croce and Matthias Hein.
\newblock Reliable evaluation of adversarial robustness with an ensemble of
  diverse parameter-free attacks.
\newblock pages 2206--2216. PMLR, 2020.

\bibitem{deng2009imagenet}
Jia Deng, Wei Dong, Richard Socher, Li-Jia Li, Kai Li, and Li Fei-Fei.
\newblock Imagenet: A large-scale hierarchical image database.
\newblock In {\em CVPR}, pages 248--255. Ieee, 2009.

\bibitem{furlanello2018born}
Tommaso Furlanello, Zachary Lipton, Michael Tschannen, Laurent Itti, and Anima
  Anandkumar.
\newblock Born again neural networks.
\newblock pages 1607--1616, 2018.

\bibitem{ge2020mutual}
Yixiao Ge, Dapeng Chen, and Hongsheng Li.
\newblock Mutual mean-teaching: Pseudo label refinery for unsupervised domain
  adaptation on person re-identification.
\newblock In {\em ICLR}, 2020.

\bibitem{ge2018fd}
Yixiao Ge, Zhuowan Li, Haiyu Zhao, Guojun Yin, Shuai Yi, Xiaogang Wang, and
  Hongsheng Li.
\newblock Fd-gan: Pose-guided feature distilling gan for robust person
  re-identification.
\newblock In {\em NeurIPS}, pages 1229--1240, 2018.

\bibitem{ge2020self}
Yixiao Ge, Haibo Wang, Feng Zhu, Rui Zhao, and Hongsheng Li.
\newblock Self-supervising fine-grained region similarities for large-scale
  image localization.
\newblock In {\em ECCV}, 2020.

\bibitem{ge2020selfpaced}
Yixiao Ge, Feng Zhu, Dapeng Chen, Rui Zhao, and Hongsheng Li.
\newblock Self-paced contrastive learning with hybrid memory for domain
  adaptive object re-id.
\newblock In {\em NeurIPS}, 2020.

\bibitem{ge2020structured}
Yixiao Ge, Feng Zhu, Rui Zhao, and Hongsheng Li.
\newblock Structured domain adaptation with online relation regularization for
  unsupervised person re-id, 2020.

\bibitem{fgsm}
Ian~J Goodfellow, Jonathon Shlens, and Christian Szegedy.
\newblock Explaining and harnessing adversarial examples.
\newblock {\em arXiv preprint arXiv:1412.6572}, 2014.

\bibitem{byol}
Jean-Bastien Grill, Florian Strub, Florent Altch\'{e}, Corentin Tallec, Pierre
  Richemond, Elena Buchatskaya, Carl Doersch, Bernardo Avila~Pires, Zhaohan
  Guo, Mohammad Gheshlaghi~Azar, Bilal Piot, koray kavukcuoglu, Remi Munos, and
  Michal Valko.
\newblock Bootstrap your own latent - a new approach to self-supervised
  learning.
\newblock In {\em NeurIPS}, volume~33, pages 21271--21284, 2020.

\bibitem{guo2020online}
Qiushan Guo, Xinjiang Wang, Yichao Wu, Zhipeng Yu, Ding Liang, Xiaolin Hu, and
  Ping Luo.
\newblock Online knowledge distillation via collaborative learning.
\newblock In {\em CVPR}, pages 11020--11029, 2020.

\bibitem{he2020momentum}
Kaiming He, Haoqi Fan, Yuxin Wu, Saining Xie, and Ross Girshick.
\newblock Momentum contrast for unsupervised visual representation learning.
\newblock In {\em CVPR}, pages 9729--9738, 2020.

\bibitem{he2017mask}
Kaiming He, Georgia Gkioxari, Piotr Doll{\'a}r, and Ross Girshick.
\newblock Mask r-cnn.
\newblock In {\em ICCV}, pages 2961--2969, 2017.

\bibitem{he2016deep}
Kaiming He, Xiangyu Zhang, Shaoqing Ren, and Jian Sun.
\newblock Deep residual learning for image recognition.
\newblock In {\em CVPR}, pages 770--778, 2016.

\bibitem{he2016identity}
Kaiming He, Xiangyu Zhang, Shaoqing Ren, and Jian Sun.
\newblock Identity mappings in deep residual networks.
\newblock In {\em ECCV}, pages 630--645. Springer, 2016.

\bibitem{hendrycks2019robustness}
Dan Hendrycks and Thomas Dietterich.
\newblock Benchmarking neural network robustness to common corruptions and
  perturbations.
\newblock {\em ICLR}, 2019.

\bibitem{hendrycks2021nae}
Dan Hendrycks, Kevin Zhao, Steven Basart, Jacob Steinhardt, and Dawn Song.
\newblock Natural adversarial examples.
\newblock 2021.

\bibitem{kd}
Geoffrey Hinton, Oriol Vinyals, and Jeffrey Dean.
\newblock Distilling the knowledge in a neural network.
\newblock In {\em Adv. Neural Inform. Process. Syst. Worksh.}, 2015.

\bibitem{huang2017densely}
Gao Huang, Zhuang Liu, Laurens Van Der~Maaten, and Kilian~Q Weinberger.
\newblock Densely connected convolutional networks.
\newblock In {\em CVPR}, pages 4700--4708, 2017.

\bibitem{pix2pix2017}
Phillip Isola, Jun-Yan Zhu, Tinghui Zhou, and Alexei~A Efros.
\newblock Image-to-image translation with conditional adversarial networks.
\newblock 2017.

\bibitem{dog}
Aditya Khosla, Nityananda Jayadevaprakash, Bangpeng Yao, and Fei-Fei Li.
\newblock Novel dataset for fine-grained image categorization: Stanford dogs.
\newblock In {\em CVPRW}, volume~2. Citeseer, 2011.

\bibitem{kim2021self}
Kyungyul Kim, ByeongMoon Ji, Doyoung Yoon, and Sangheum Hwang.
\newblock Self-knowledge distillation with progressive refinement of targets.
\newblock In {\em ICCV}, pages 6567--6576, 2021.

\bibitem{cifar}
Alex Krizhevsky, Geoffrey Hinton, et~al.
\newblock Learning multiple layers of features from tiny images.
\newblock 2009.

\bibitem{temensemble}
Samuli Laine and Timo Aila.
\newblock Temporal ensembling for semi-supervised learning.
\newblock In {\em ICLR}, 2017.

\bibitem{lan2018one}
Xu Lan, Xiatian Zhu, and Shaogang Gong.
\newblock Knowledge distillation by on-the-fly native ensemble.
\newblock In {\em NeurIPS}, volume~31, 2018.

\bibitem{lin2017feature}
Tsung-Yi Lin, Piotr Doll{\'a}r, Ross Girshick, Kaiming He, Bharath Hariharan,
  and Serge Belongie.
\newblock Feature pyramid networks for object detection.
\newblock In {\em CVPR}, pages 2117--2125, 2017.

\bibitem{coco}
Tsung-Yi Lin, Michael Maire, Serge Belongie, James Hays, Pietro Perona, Deva
  Ramanan, Piotr Doll{\'a}r, and C~Lawrence Zitnick.
\newblock Microsoft coco: Common objects in context.
\newblock In {\em ECCV}, pages 740--755. Springer, 2014.

\bibitem{Liu_DivCo}
Rui Liu, Yixiao Ge, Ching~Lam Choi, Xiaogang Wang, and Hongsheng Li.
\newblock Divco: Diverse conditional image synthesis via contrastive generative
  adversarial network.
\newblock In {\em CVPR}, 2021.

\bibitem{liu2021swin}
Ze Liu, Yutong Lin, Yue Cao, Han Hu, Yixuan Wei, Zheng Zhang, Stephen Lin, and
  Baining Guo.
\newblock Swin transformer: Hierarchical vision transformer using shifted
  windows.
\newblock In {\em ICCV}, 2021.

\bibitem{mahajan2018exploring}
Dhruv Mahajan, Ross Girshick, Vignesh Ramanathan, Kaiming He, Manohar Paluri,
  Yixuan Li, Ashwin Bharambe, and Laurens Van Der~Maaten.
\newblock Exploring the limits of weakly supervised pretraining.
\newblock In {\em ECCV}, pages 181--196, 2018.

\bibitem{park2019relational}
Wonpyo Park, Dongju Kim, Yan Lu, and Minsu Cho.
\newblock Relational knowledge distillation.
\newblock In {\em CVPR}, pages 3967--3976, 2019.

\bibitem{mit}
Ariadna Quattoni and Antonio Torralba.
\newblock Recognizing indoor scenes.
\newblock In {\em CVPR}, pages 413--420. IEEE, 2009.

\bibitem{radosavovic2020designing}
Ilija Radosavovic, Raj~Prateek Kosaraju, Ross Girshick, Kaiming He, and Piotr
  Doll{\'a}r.
\newblock Designing network design spaces.
\newblock In {\em CVPR}, pages 10428--10436, 2020.

\bibitem{fasterrcnn}
Shaoqing Ren, Kaiming He, Ross Girshick, and Jian Sun.
\newblock Faster r-cnn: Towards real-time object detection with region proposal
  networks.
\newblock In {\em NeurIPS}, volume~28, 2015.

\bibitem{howard2017mobilenets}
Mark Sandler, Andrew Howard, Menglong Zhu, Andrey Zhmoginov, and Liang-Chieh
  Chen.
\newblock Mobilenetv2: Inverted residuals and linear bottlenecks.
\newblock In {\em CVPR}, June 2018.

\bibitem{shankar2020evaluating}
Vaishaal Shankar, Rebecca Roelofs, Horia Mania, Alex Fang, Benjamin Recht, and
  Ludwig Schmidt.
\newblock Evaluating machine accuracy on imagenet.
\newblock pages 8634--8644. PMLR, 2020.

\bibitem{shen2019meal}
Zhiqiang Shen, Zhankui He, and Xiangyang Xue.
\newblock Meal: Multi-model ensemble via adversarial learning.
\newblock In {\em AAAI}, volume~33, pages 4886--4893, 2019.

\bibitem{son2021densely}
Wonchul Son, Jaemin Na, Junyong Choi, and Wonjun Hwang.
\newblock Densely guided knowledge distillation using multiple teacher
  assistants.
\newblock In {\em ICCV}, pages 9395--9404, 2021.

\bibitem{sun2017revisiting}
Chen Sun, Abhinav Shrivastava, Saurabh Singh, and Abhinav Gupta.
\newblock Revisiting unreasonable effectiveness of data in deep learning era.
\newblock In {\em ICCV}, pages 843--852, 2017.

\bibitem{szegedy2015going}
Christian Szegedy, Wei Liu, Yangqing Jia, Pierre Sermanet, Scott Reed, Dragomir
  Anguelov, Dumitru Erhan, Vincent Vanhoucke, and Andrew Rabinovich.
\newblock Going deeper with convolutions.
\newblock In {\em CVPR}, pages 1--9, 2015.

\bibitem{szegedy2016rethinking}
Christian Szegedy, Vincent Vanhoucke, Sergey Ioffe, Jon Shlens, and Zbigniew
  Wojna.
\newblock Rethinking the inception architecture for computer vision.
\newblock In {\em CVPR}, pages 2818--2826, 2016.

\bibitem{tan2019efficientnet}
Mingxing Tan and Quoc Le.
\newblock Efficientnet: Rethinking model scaling for convolutional neural
  networks.
\newblock pages 6105--6114, 2019.

\bibitem{meanteacher}
Antti Tarvainen and Harri Valpola.
\newblock Mean teachers are better role models: Weight-averaged consistency
  targets improve semi-supervised deep learning results.
\newblock In {\em NeurIPS}, volume~30, 2017.

\bibitem{tian2019crd}
Yonglong Tian, Dilip Krishnan, and Phillip Isola.
\newblock Contrastive representation distillation.
\newblock In {\em ICLR}, 2020.

\bibitem{tung2019similarity}
Frederick Tung and Greg Mori.
\newblock Similarity-preserving knowledge distillation.
\newblock In {\em ICCV}, pages 1365--1374, 2019.

\bibitem{cub}
Catherine Wah, Steve Branson, Peter Welinder, Pietro Perona, and Serge
  Belongie.
\newblock The caltech-ucsd birds-200-2011 dataset.
\newblock 2011.

\bibitem{xie2020self}
Qizhe Xie, Minh-Thang Luong, Eduard Hovy, and Quoc~V Le.
\newblock Self-training with noisy student improves imagenet classification.
\newblock In {\em CVPR}, pages 10687--10698, 2020.

\bibitem{Xie2016}
Saining Xie, Ross Girshick, Piotr Dollar, Zhuowen Tu, and Kaiming He.
\newblock Aggregated residual transformations for deep neural networks.
\newblock In {\em CVPR}, July 2017.

\bibitem{xu2019data}
Ting-Bing Xu and Cheng-Lin Liu.
\newblock Data-distortion guided self-distillation for deep neural networks.
\newblock In {\em AAAI}, volume~33, pages 5565--5572, 2019.

\bibitem{yuan2020revisiting}
Li Yuan, Francis~EH Tay, Guilin Li, Tao Wang, and Jiashi Feng.
\newblock Revisiting knowledge distillation via label smoothing regularization.
\newblock In {\em CVPR}, pages 3903--3911, 2020.

\bibitem{yun2019cutmix}
Sangdoo Yun, Dongyoon Han, Seong~Joon Oh, Sanghyuk Chun, Junsuk Choe, and
  Youngjoon Yoo.
\newblock Cutmix: Regularization strategy to train strong classifiers with
  localizable features.
\newblock In {\em ICCV}, pages 6023--6032, 2019.

\bibitem{yun2021relabel}
Sangdoo Yun, Seong~Joon Oh, Byeongho Heo, Dongyoon Han, Junsuk Choe, and
  Sanghyuk Chun.
\newblock Re-labeling imagenet: from single to multi-labels, from global to
  localized labels.
\newblock In {\em CVPR}, 2021.

\bibitem{yun2020regularizing}
Sukmin Yun, Jongjin Park, Kimin Lee, and Jinwoo Shin.
\newblock Regularizing class-wise predictions via self-knowledge distillation.
\newblock In {\em CVPR}, pages 13876--13885, 2020.

\bibitem{zhang2020resnest}
Hang Zhang, Chongruo Wu, Zhongyue Zhang, Yi Zhu, Zhi Zhang, Haibin Lin, Yue
  Sun, Tong He, Jonas Muller, R. Manmatha, Mu Li, and Alexander Smola.
\newblock Resnest: Split-attention networks.
\newblock {\em arXiv preprint arXiv:2004.08955}, 2020.

\bibitem{zhang2019your}
Linfeng Zhang, Jiebo Song, Anni Gao, Jingwei Chen, Chenglong Bao, and Kaisheng
  Ma.
\newblock Be your own teacher: Improve the performance of convolutional neural
  networks via self distillation.
\newblock In {\em ICCV}, pages 3713--3722, 2019.

\bibitem{zhang2018deep}
Ying Zhang, Tao Xiang, Timothy~M Hospedales, and Huchuan Lu.
\newblock Deep mutual learning.
\newblock In {\em CVPR}, pages 4320--4328, 2018.

\bibitem{zhao2017pyramid}
Hengshuang Zhao, Jianping Shi, Xiaojuan Qi, Xiaogang Wang, and Jiaya Jia.
\newblock Pyramid scene parsing network.
\newblock In {\em CVPR}, pages 2881--2890, 2017.

\bibitem{zhou2003learning}
Dengyong Zhou, Olivier Bousquet, Thomas Lal, Jason Weston, and Bernhard
  Sch{\"o}lkopf.
\newblock Learning with local and global consistency.
\newblock {\em NeurIPS}, 16, 2003.

\bibitem{CycleGAN2017}
Jun-Yan Zhu, Taesung Park, Phillip Isola, and Alexei~A Efros.
\newblock Unpaired image-to-image translation using cycle-consistent
  adversarial networkss.
\newblock In {\em ICCV}, 2017.

\end{thebibliography}
}

\end{document}